%% file: main.tex
\documentclass[10pt,twocolumn,letterpaper]{article}

\usepackage{cvpr}              
\input{preamble}
\definecolor{cvprblue}{rgb}{0.21,0.49,0.74}
\usepackage[pagebackref,breaklinks,colorlinks,allcolors=cvprblue]{hyperref}

\usepackage{graphicx}
\usepackage{amsmath}

\usepackage{multirow}
\usepackage{rotating}
\usepackage{array}
\usepackage{bbding}
\usepackage{wrapfig}  
\usepackage[accsupp]{axessibility}

\def\paperID{4533} 
\def\confName{CVPR}
\def\confYear{2026}

\title{Spectral Scalpel: Amplifying Adjacent Action Discrepancy via Frequency-Selective Filtering for Skeleton-Based Action Segmentation}

\author{
    Haoyu Ji\textsuperscript{1}, 
    Bowen Chen\textsuperscript{2}, 
    Zhihao Yang\textsuperscript{1}, 
    Wenze Huang\textsuperscript{1}, 
    Yu Gao\textsuperscript{1},
    Xueting Liu\textsuperscript{3},\\
    Weihong Ren\textsuperscript{1},
    Zhiyong Wang\textsuperscript{1,*},
    Honghai Liu\textsuperscript{1,4}\\[2mm]
    \textsuperscript{1}Harbin Institute of Technology, Shenzhen \quad
    \textsuperscript{2}Shenzhen H\&T Intelligent Control Co., Ltd. \\
    \textsuperscript{3}Southern University of Science and Technology  \quad
    \textsuperscript{4}Southeast University \\
}

\begin{document}
\maketitle

{
\let\thefootnote\relax\footnotetext{$^*$Corresponding author}
}

\input{sec/0_abstract}    
\input{sec/1_intro}
\input{sec/2_relatedwork}
\input{sec/3_method}
\input{sec/4_experiment}

\input{sec/5_conclusion}

\section*{Acknowledgement}
This work was supported in part by the National Key Research and Development Program of China under Grant 2025YFE0218500; in part by the National Natural Science Foundation of China under Grant 62503139, Grant 62573163, and Grant 62521006; in part by Shenzhen Science and Technology Program under Grant JCYJ20240813105137049; and in part by the GuangDong Basic and Applied Basic Research Foundation under Grant 2024A1515012028, and Grant 2026A1515011822; and in part by the Shenzhen Medical Research Fund under Grant A2502034.

{
    \small
    \bibliographystyle{ieeenat_fullname}
    \bibliography{main}
}

\input{sec/X_suppl}

\end{document}

%% file: sec/0_abstract.tex
\begin{abstract}
Skeleton-based Temporal Action Segmentation (STAS) seeks to densely segment and classify diverse actions within long, untrimmed skeletal motion sequences. However, existing STAS methodologies face challenges of limited inter-class discriminability and blurred segmentation boundaries, primarily due to insufficient distinction of spatio-temporal patterns between adjacent actions. To address these limitations, we propose Spectral Scalpel, a frequency-selective filtering framework aimed at suppressing shared frequency components between adjacent distinct actions while amplifying their action-specific frequencies, thereby enhancing inter-action discrepancies and sharpening transition boundaries. Specifically, Spectral Scalpel employs adaptive multi-scale spectral filters as scalpels to edit frequency spectra, coupled with a discrepancy loss between adjacent actions serving as the surgical objective. This design amplifies representational disparities between neighboring actions, effectively mitigating boundary localization ambiguities and inter-class confusion. Furthermore, complementing long-term temporal modeling, we introduce a frequency-aware channel mixer to strengthen channel evolution by aggregating spectra across channels. This work presents a novel paradigm for STAS that extends conventional spatio-temporal modeling by incorporating frequency-domain analysis. Extensive experiments on five public datasets demonstrate that Spectral Scalpel achieves state-of-the-art performance. Code is available at \url{https://github.com/HaoyuJi/SpecScalpel}.
\end{abstract}

%% file: sec/1_intro.tex
\section{Introduction}
\label{sec:intro}

\begin{figure}[t]
    \centering
    \includegraphics[width=1.0\linewidth]{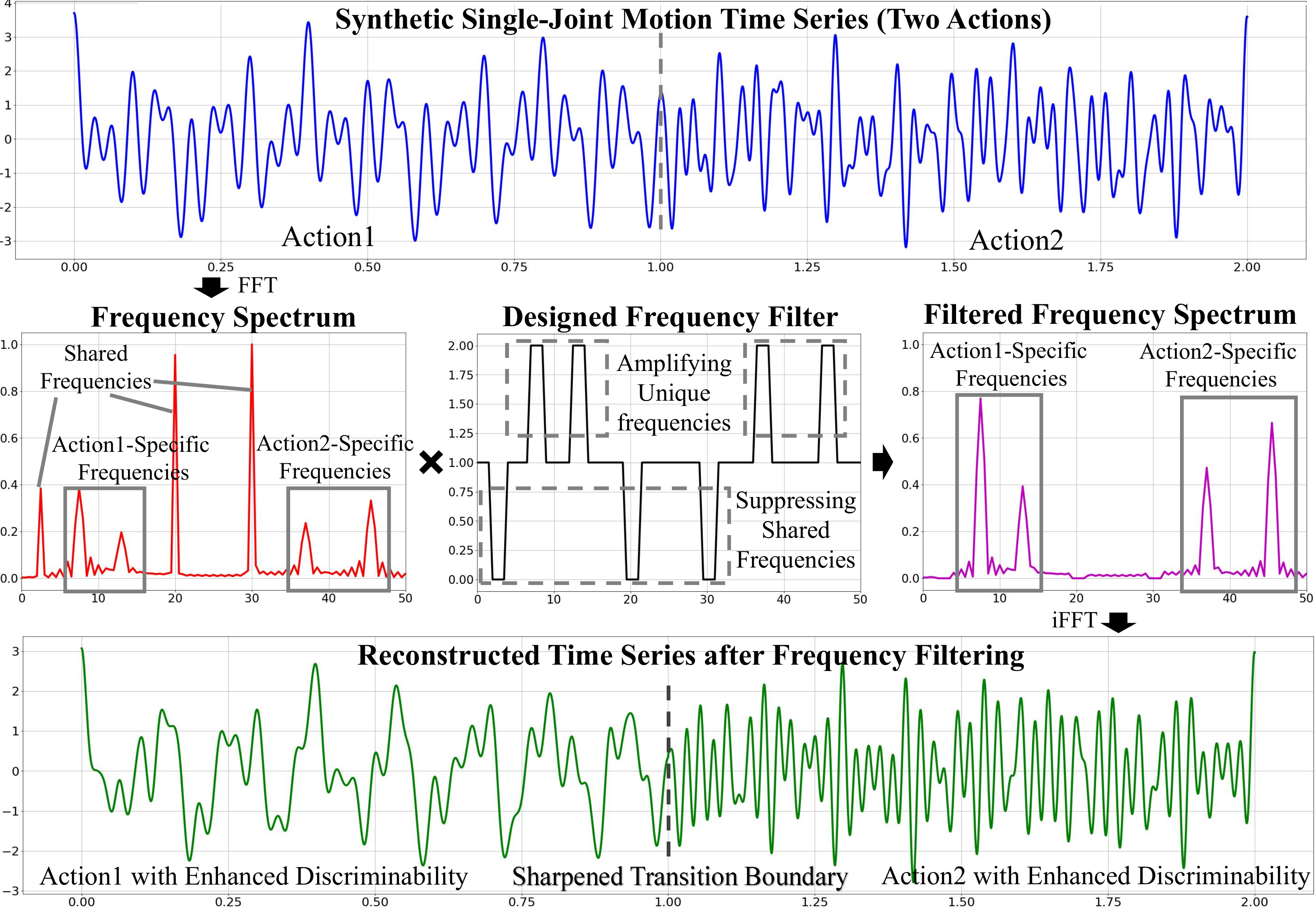}
    \caption{Effect of a designed frequency filter in amplifying adjacent action discrepancy. The synthetic single-joint motion sequence consists of two concatenated action segments, each idealized as a sum of cosine waves with both shared and action-specific frequency components. After FFT, a designed filter amplifies action-specific frequencies and suppresses shared ones. The filtered sequence shows enhanced action discriminability and a clearer boundary.}
    \label{fig:1}
\end{figure}

Temporal Action Segmentation (TAS) is a high-level task in video understanding, with broad applications in medical rehabilitation~\cite{AutoENP,Auto_freezing}, industrial monitoring~\cite{HA-ViD,Openpack}, sports analysis~\cite{3D_skating,MCFS}, and human-computer interaction~\cite{HCI}. The objective of TAS is to assign action labels to each frame in an untrimmed human activity video~\cite{TAS}. Skeleton-based Temporal Action Segmentation (STAS) parses human motion through skeletal sequences, offering a lightweight and robust alternative to Video-based approaches (VTAS)~\cite{IDT-GCN,LaSA}. Existing STAS methods have made notable progress by leveraging spatio-temporal modeling~\cite{IDT-GCN,LaSA, MS-GCN, SFA, DeST}, which captures spatial dependencies among joints and temporal correlations across frames.

Current STAS approaches, however, still suffer from two critical limitations: 1) Inter-class confusion arising from insufficiently discriminative features among visually or semantically similar action categories, and 2) Boundary ambiguity caused by indistinguishable transition dynamics between adjacent actions with high similarity. These issues stem from a fundamental dilemma in existing temporal modeling paradigms (e.g., TCNs and Transformers). By essentially acting as low-pass filters to aggregate context, these models introduce an inherent smoothing effect. While highly effective at maintaining intra-action coherence, this over-smoothing inadvertently obscures the essential discrepancies between adjacent distinct actions and blurs sharp transition boundaries. Consequently, there is an urgent need to explicitly capture and amplify the core differences to compensate for this architectural deficiency. 

To break this bottleneck, we pivot our perspective to the frequency domain. Fundamentally, skeletal motions are composed of joint oscillations, meaning distinct actions exhibit highly divergent periodicities and spectral energy distributions rather than merely superficial temporal trajectories. Specifically, the frequency spectra of different actions inevitably contain both shared common components and unique, action-specific characteristics. Because these subtle action-specific cues are easily erased by the smoothing effect of temporal modeling, it is imperative to explicitly amplify them from a frequency perspective. As illustrated in Fig.~\ref{fig:1}, selectively filtering the motion spectrum to suppress shared components and enhance action-specific frequencies effectively amplifies inter-class discriminability and sharpens transition boundaries.

\begin{figure}[t]
    \centering
    \includegraphics[width=0.9\linewidth]{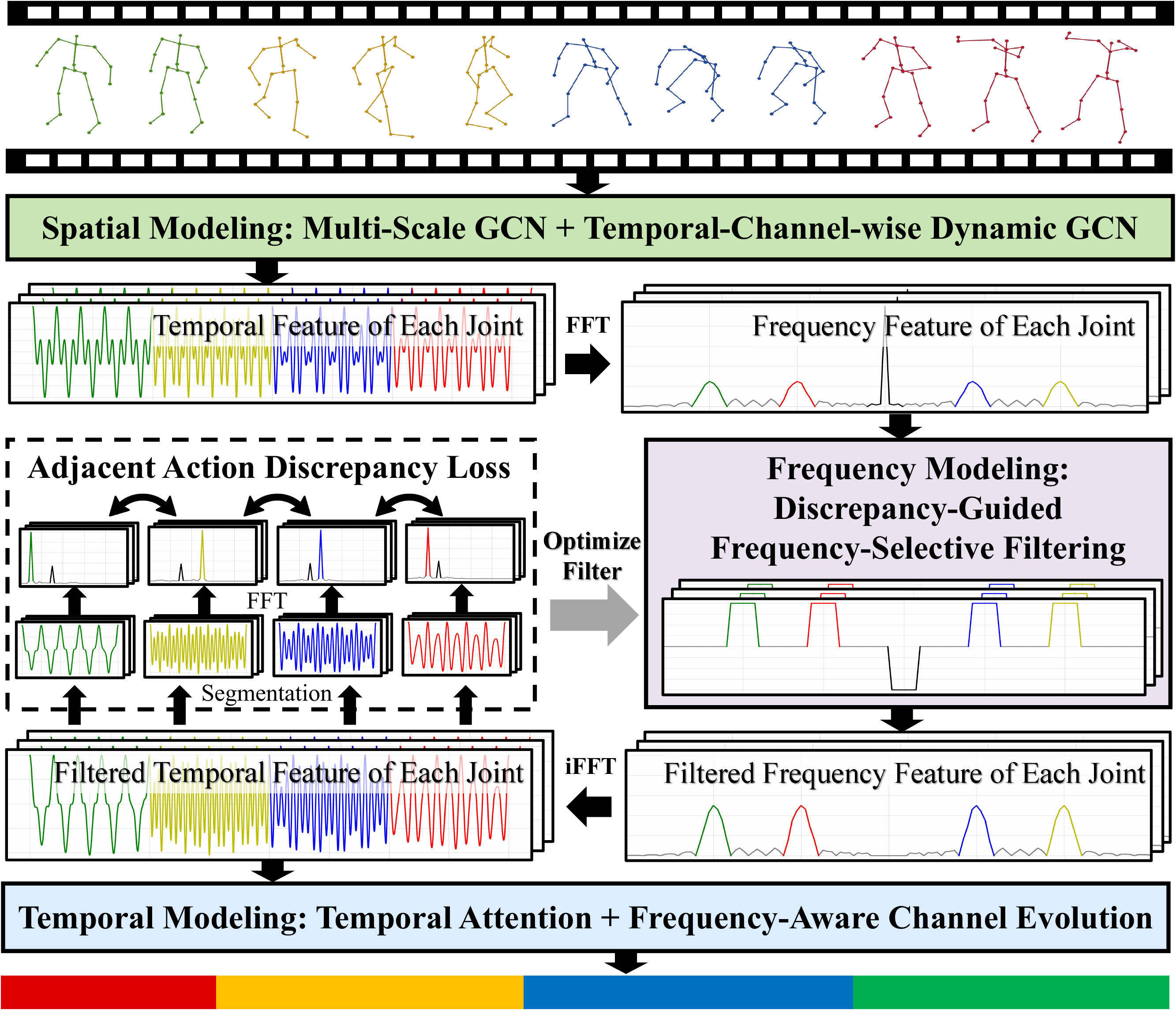}
    \caption{Overview of Spectral Scalpel. After spatial modeling, joint features are transformed into the frequency domain. A discrepancy-guided filter, optimized by adjacent action discrepancy loss, enhances action-specific spectra and suppresses shared ones, improving discriminability and boundary clarity.}
    \label{fig:2}
\end{figure}

Motivated by this insight, we propose \textbf{Spectral Scalpel}, a novel discrepancy-guided frequency-selective filtering framework for STAS. Unlike prior Fast Fourier Transform (FFT)-based filtering methods~\cite{AFF, DFFormer} that rely on unconstrained, blind data-driven learning, our approach performs active spectral editing guided by explicit objectives. 
Analogous to a surgical procedure, where the scalpel serves the surgical objective, our method treats the amplification of adjacent action differences as the ``objective'' and employs adaptive filtering as the ``scalpel'' to sharpen feature distinctions, as shown in Fig.~\ref{fig:2}.

Specifically, the Multi-scale Adaptive Spectral Filter (MASF), functioning as the scalpel, transforms temporal features into the frequency domain via FFT, applies learnable filters at multiple scales, and fuses the filtered features via a dual-branch dynamic-static channel-wise weighting scheme. 
The Adjacent Action Discrepancy Loss (AADL), serving as the surgical objective, segments filtered features into action-specific units, transforms them into the frequency domain, and explicitly maximizes the spectral discrepancy between adjacent action segments.
Additionally, to strengthen subsequent temporal modeling, we introduce the Frequency-Aware Channel Mixer (FACM) to capture frequency-sensitive interactions across feature channels directly in the spectral space.
Extensive experiments on five challenging STAS datasets---PKU-MMD v2 (X-sub)~\cite{PKU-MMD}, PKU-MMD v2 (X-view)~\cite{PKU-MMD}, LARa~\cite{LARA}, MCFS-130~\cite{MCFS}, and TCG-15~\cite{tcg}---demonstrate that Spectral Scalpel achieves state-of-the-art performance while maintaining competitive computational efficiency.

The main contributions are summarized as follows:
1) We propose Spectral Scalpel, the first framework to systematically integrate frequency-domain processing into STAS, enhancing inter-class discriminability and sharpening transition boundaries.
2) We design a discrepancy-guided filtering mechanism comprising the Multi-scale Adaptive Spectral Filter (MASF) and Adjacent Action Discrepancy Loss (AADL), which optimizes spectral discrepancy between actions to effectively guide the filter in learning discriminative representations.
3) We introduce a Frequency-Aware Channel Mixer (FACM) module that strengthens temporal modeling through spectral interaction among channels, enabling frequency-domain-aware channel evolution.

%% file: sec/2_relatedwork.tex
\section{Related Work}
\label{sec:related_work}

\noindent\textbf{Temporal Action Segmentation.}
Video-based Temporal Action Segmentation (VTAS) processes RGB or optical flow sequences through temporal modeling. Temporal Convolutional Networks (TCNs)~\cite{TCN, MS-TCN, ETSN} became popular for modeling long-term dependencies. Subsequent methods enhanced TCNs via dual dilated layers~\cite{MS-TCN++}, receptive field design~\cite{Global2local,RF-Next}, multi-scale fusion~\cite{C2F-TCN}, boundary awareness~\cite{BCN,ASRF}, dilated propagation and reconstruction~\cite{DPRN}, and diffusion modeling~\cite{Diffusion}. Some graph-based approaches~\cite{GTRM,DTGRM,Semantic2Graph} model temporal semantics via graph convolutions. Transformer-based methods improve temporal modeling via ASFormer~\cite{Asformer}, seq2seq translation~\cite{UVAST}, U-shaped designs~\cite{EUT}, sparse attention~\cite{LTContext}, boundary-aware voting~\cite{BaFormer}, and frame-action attention~\cite{FACT}.
Skeleton-based Temporal Action Segmentation (STAS) leverages spatio-temporal modeling over skeleton sequences. MS-GCN~\cite{MS-GCN} combines Graph Convolutional Networks (GCNs) for spatial modeling and TCNs for temporal modeling. Subsequent methods further enhanced spatio-temporal capabilities, such as discriminative temporal graph networks~\cite{IDT-GCN}, decoupled spatio-temporal frameworks~\cite{DeST}, motion-aware and temporally enhanced methods~\cite{MTST-GCN}, contrastive learning-based representation optimization~\cite{SCSAS}, and multi-scale feature pyramids~\cite{MSTAS}. Attention-based approaches further enhance spatial-temporal focus, such as spatial focus attention~\cite{SFA}, spatio-temporal graph attention~\cite{STGA-Net}, and snippet-aware Transformer with multiple action elements~\cite{ME-ST}. Other methods focus on data processing strategies~\cite{Tai_Chi, LAC, CTC}. Additionally, language-assisted methods~\cite{LaSA, LPL, TRG-Net} use textual cues~\cite{BERT} to enhance feature modeling. Spectral Scalpel is the first to explore frequency-domain modeling in STAS, leveraging discrepancy-guided spectral filters to enhance feature discriminability.

\noindent\textbf{FFT-based Frequency Domain Analysis Methods.}
Frequency-domain analysis has become a powerful tool in computer vision, enabling efficient global modeling and uncovering periodic patterns beyond the reach of spatial-temporal methods. Fast Fourier Transform (FFT) is widely adopted for its computational efficiency. Notable methods include FFC~\cite{FFC} for global feature mapping, AFNO~\cite{AFNO} for adaptive token mixing, AFF~\cite{AFF} for semantic-aware filtering, and DFFormer~\cite{DFFormer} with dynamic frequency filters. In image restoration, FFT-based techniques such as frequency self-attention and gating~\cite{FSAS}, Fourier spatial interaction and channel evolution~\cite{Fourmer}, dynamic aggregation with spatial-frequency interaction~\cite{SFSNiD}, and Fourier convolution with contrastive regularization~\cite{FADformer} have shown great promise. For segmentation and detection, frequency-aware fusion improves consistency and boundary sharpness~\cite{FreqFusion}. In video understanding, FAR~\cite{FAR} introduces Fourier-based object disentanglement, while DFN~\cite{DFN} employs a sliding-window Fourier token mixer for action segmentation. Inspired by these advances, Spectral Scalpel explores frequency-domain modeling for skeletal motion, using discrepancy-guided spectral filtering to amplify adjacent action distinctions and improve feature discrimination and boundary precision.

%% file: sec/3_method.tex
\section{Method}
\label{sec:method}

\begin{figure*}[t]
  \centering
   \includegraphics[width=0.95\linewidth]{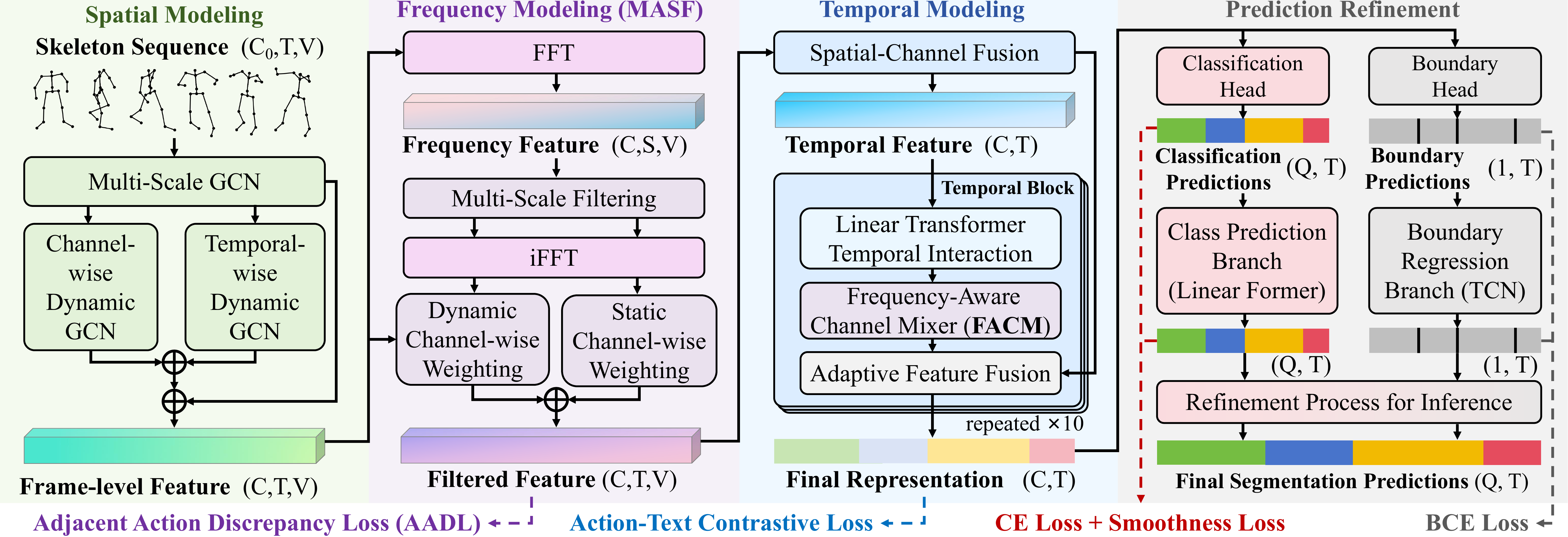}

   \caption{Overview of the Spectral Scalpel framework. It comprises four sequential stages: (1) Spatial Modeling with multi-scale, dual-branch dynamic GCNs; (2) Frequency Modeling via the Multi-scale Adaptive Spectral Filter (MASF); (3) Temporal Modeling through Linear Transformers, Frequency-Aware Channel Mixers (FACM), and adaptive fusion; (4) Prediction Refinement, which generates and refines class and boundary predictions. The model is optimized by supervised losses computed from filtered features, final representations, classification predictions, and boundary predictions.}
   \label{fig:3}
\end{figure*}

STAS processes skeleton sequences denoted as \( X_{1:T} = [x_1, x_2, \dots, x_T] \in \mathbb{R}^{C_0 \times T \times V} \), where \( x_t \in \mathbb{R}^{C_0 \times V} \) represents the \( t \)-th frame, with \( V \) and \( C_0 \) denoting the number of joints and channels, respectively. STAS maps the input \( X_{1:T} \) to the label space defined by the class set \( Q \) as \( Y_{1:T} = [y_1, y_2, \dots, y_T] \in \mathbb{R}^{Q \times T} \), where \( y_t \in \{0,1\}^Q \) indicates the one-hot action label for the \( t \)-th frame.

\subsection{Overall Framework}

The overall framework of Spectral Scalpel is illustrated in Fig.~\ref{fig:3}. Given a skeleton sequence $X \in \mathbb{R}^{C_0 \times T \times V}$ as input, the model sequentially performs spatial modeling, frequency modeling, and temporal modeling to obtain the final representation $F_R \in \mathbb{R}^{C \times T}$. The representation $F_R$ is then fed into a classification head and a boundary head to produce class and boundary predictions, which are subsequently refined. The model is optimized by minimizing a loss function computed from multiple feature representations and predictions.

\textbf{Spatial Modeling}:
The spatial modeling follows the prior STAS framework~\cite{TRG-Net}. Given the input sequence $X \in \mathbb{R}^{C_0 \times T \times V}$, a multi-scale GCN~\cite{DeST, MS-G3D} leveraging the prior topological graph of human joints is applied to produce initial spatial features $F_{s0} \in \mathbb{R}^{C \times T \times V}$. Then, dynamic GCNs~\cite{MTST-GCN, TRG-Net, CTR-GCN} along the channel and temporal dimensions are used to adaptively model fine-grained joint spatial relationships per channel and per frame, resulting in refined spatial features $F_s \in \mathbb{R}^{C \times T \times V}$.

\textbf{Frequency Modeling}:
The core innovation of Spectral Scalpel lies in its frequency modeling, termed Multi-scale Adaptive Spectral Filter (MASF). An FFT is first applied to $F_s$ along the temporal axis, converting the signal to the frequency domain as $F_{f0} \in \mathbb{C}^{C \times S \times V}$. Then, multi-scale multi-head spectral filters are used to perform frequency-domain filtering. These filtered signals are transformed back to the time domain via inverse FFT, producing $F_f^1, \dots, F_f^M \in \mathbb{R}^{C \times T \times V}$. These are subsequently fused through dual dynamic-static channel-wise weighting branches, resulting in the filtered feature $F_f \in \mathbb{R}^{C \times T \times V}$. Notably, during training, $F_{f}$ is enhanced via an Adjacent Action Discrepancy Loss (AADL) to improve action discriminability and boundary clarity.

\textbf{Temporal Modeling}:
The temporal modeling largely follows the STAS pipeline~\cite{TRG-Net}. Firstly, the spatial-channel fusion compresses the channels of $F_{f}$, and integrates the spatial dimension into the channel dimension through convolution, yielding the temporal feature \(F_{t0}^0\in \mathbb{R}^{C \times T}\). This feature is then processed by a stack of temporal blocks. Each block integrates global temporal dependencies via a Linear Transformer~\cite{Linformer1, Linformer2} and enhances channel evolution from a spectral perspective using the proposed Frequency-Aware Channel Mixer (FACM). Similar to~\cite{DeST, TRG-Net}, we employ adaptive feature fusion to preserve multi-level core features. Specifically, the temporally modeled features from each block $F_{t2}^l \in \mathbb{R}^{C \times T}$ are fused with the early-stage temporal features $F_{t0}^l \in \mathbb{R}^{C \times T}$ extracted by another spatial-channel fusion head, producing $F_t^l \in \mathbb{R}^{C \times T}$. The output of the final block serves as the final representation $F_R \in \mathbb{R}^{C \times T}$, which is further guided during training via action-text contrastive learning~\cite{LaSA, TRG-Net} to improve semantic quality.

\textbf{Prediction Refinement}:
The prediction refinement follows the TAS design~\cite{LaSA, DeST, ASRF, TRG-Net}. The representation $F_R$ is fed into a classification head and a boundary regression head to produce initial class and boundary predictions, $Y^c_0 \in \mathbb{R}^{Q \times T}$ and $Y^b_0 \in \mathbb{R}^{1 \times T}$, respectively. For classification, $Y^c_0$ is refined via a branch of $S^c$ stages, each comprising 10 layers of Linear Transformers for cross-attention, yielding the final class prediction $Y^c_F \in \mathbb{R}^{Q \times T}$. For boundary regression, $Y^b_0$ is refined through $S^b$ stages of 10-layer Dilated TCNs, resulting in the final boundary prediction $Y^b_F \in \mathbb{R}^{1 \times T}$. During training, all stages of $Y^c$ are supervised via cross-entropy and smoothing losses, while all $Y^b$ are trained using binary cross-entropy loss. During inference, the final segmentation result is obtained by jointly considering $Y^c_F$ and $Y^b_F$ following the ASRF post-processing scheme~\cite{ASRF}.

\begin{figure*}[t]
  \centering
   \includegraphics[width=0.98\linewidth]{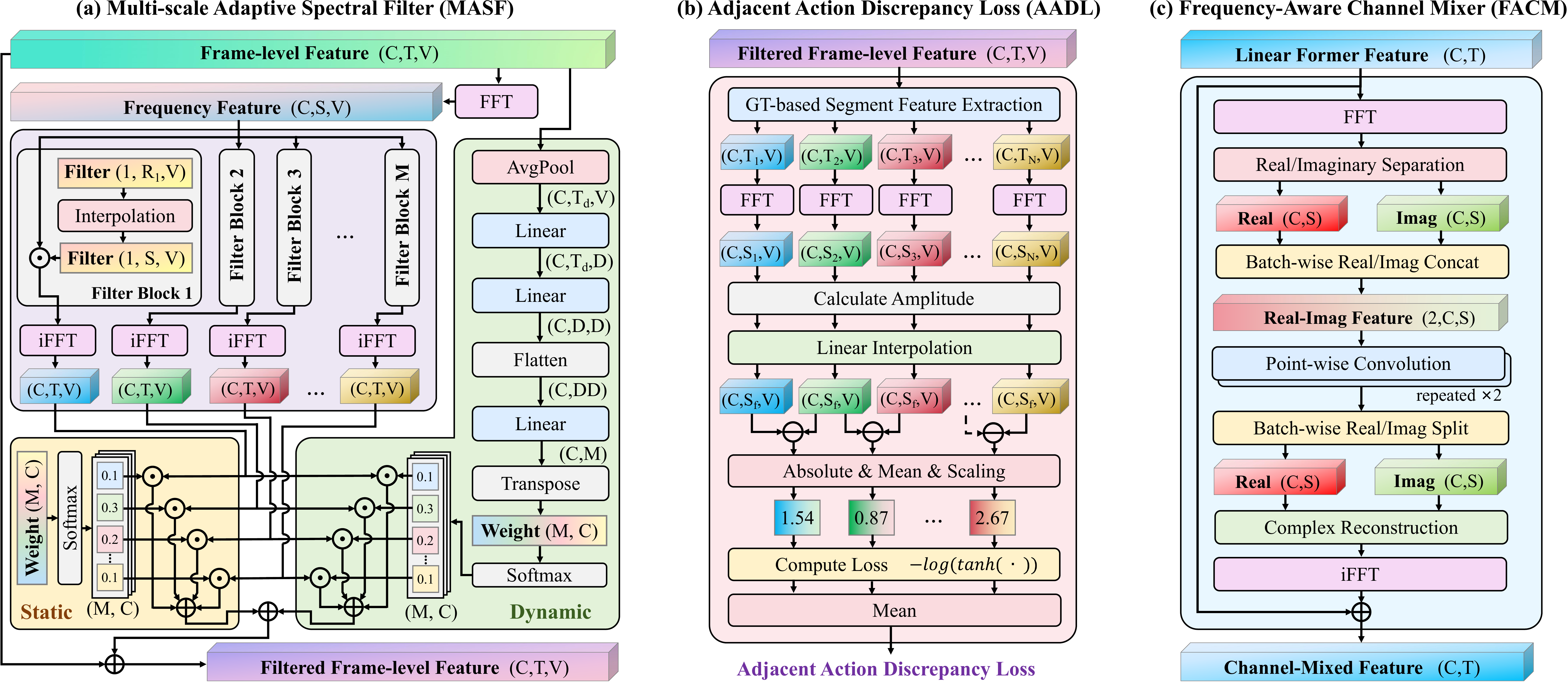}

   \caption{Illustration of the three core components. (a) Multi-scale Adaptive Spectral Filter (MASF) enhances frequency discriminability using multi-head filters at multiple scales, combined with dual-branch dynamic-static channel-wise fusion. (b) Adjacent Action Discrepancy Loss (AADL) guides frequency dynamics by maximizing amplitude spectrum discrepancies between adjacent segments. (c) Frequency-Aware Channel Mixer (FACM) models frequency-domain channel interactions via real-imaginary decomposition and shared point-wise convolutions.}
   \label{fig:4}
\end{figure*}

\subsection{Multi-scale Adaptive Spectral Filter}

As the ``scalpel'' for enhancing frequency-specific discriminability, the Multi-scale Adaptive Spectral Filter (MASF) is designed to conduct scale-specific spectral filtering through multiple heads, followed by adaptive fusion. As illustrated in Fig.~\ref{fig:4}(a), MASF comprises two main stages: (i) Multi-scale Multi-head Filtering, and (ii) Dual-branch Dynamic-Static Channel-wise Fusion.

\textbf{Multi-scale Multi-head Filtering.}
Given spatial features $F_s$, an FFT is applied along the temporal axis to obtain the initial frequency feature $F_{f0} \in \mathbb{C}^{C \times S \times V}$. MASF then applies multi-scale filtering through $M$ filter blocks. Each block contains a learnable spectral filter $H_m \in \mathbb{R}^{1 \times R_m \times V}$, where $R_m$ denotes the filter length or scale. The filter is interpolated to the spectral resolution $S$ using Nearest Neighbor Interpolation (NNI), and applied via Hadamard (element-wise) product with the frequency representation $F_{f0}$. The filtered result is transformed back to the time domain using an inverse FFT. The filtered feature of the $m$-th block $F_f^m \in \mathbb{R}^{C \times T \times V }$ is calculated as follows:
\begin{equation}
F_f^m = \mathcal{F}^{-1} \left( \mathrm{NNI}(H_m) \odot \mathcal{F}(F_s) \right),
\end{equation}
where $\mathcal{F}$ and $\mathcal{F}^{-1}$ denote the FFT and inverse FFT operations, respectively, and $\odot$ is the Hadamard product. All filters $H_m$ are initialized with ones, allowing the model to learn to suppress or amplify specific and shared frequencies through training.
To capture frequency components at various resolutions and scales, each filter uses a distinct length $R_m$, determined as:
\begin{equation}
R_m = \frac{(M + m) \cdot R_{\text{max}}}{2M},
\label{equ:2}
\end{equation}
where $R_{\text{max}}$ is the maximum filter length. Unlike exponential strategies such as $R_m = 2^{m + b}$ (with constant $b$), the proposed linear formulation ensures less overlap between the interpolated filter boundaries, thus facilitating finer-grained and more distinctive channel-wise weighting.

\textbf{Dual-branch Dynamic-Static Channel-wise Fusion.}
To aggregate the multi-scale filtered features $[F_f^1, F_f^2, \dots, F_f^M]$, we propose a dual-branch fusion strategy that performs channel-wise weighting through complementary static and dynamic branches, achieving both generalizability and instance-specific adaptability.
The static branch employs a learnable parameter matrix $W_{st} \in \mathbb{R}^{M \times C}$ that remains fixed after training, establishing cross-sample consistent but inflexible patterns. In contrast, the dynamic branch generates input-conditioned weights $W_{dy} \in \mathbb{R}^{M \times C}$ via learnable linear transformations of $F_s$, providing highly flexible but sample-dependent mappings:
\begin{equation}
W_{dy} = \left[ W_M \cdot \mathrm{Flatten}(W_T \cdot W_S \cdot \mathrm{AvgPool}(F_s)) \right]^\top,
\label{equ:3}
\end{equation}
where $\mathrm{AvgPool}$ adaptively downsamples the temporal dimension to a fixed length $T_d$, and $W_S \in \mathbb{R}^{V \times D}$, $W_T \in \mathbb{R}^{T_d \times D}$, and $W_M \in \mathbb{R}^{DD \times M}$ are trainable projection matrices. The $\mathrm{Flatten}$ operator reshapes the intermediate feature from $\mathbb{R}^{C \times D \times D}$ to $\mathbb{R}^{C \times DD}$.
Finally, we apply a softmax normalization along the dimension $M$ for both static and dynamic weights, and compute the dual-branch channel-wise fused feature $F_f \in \mathbb{R}^{C \times T \times V}$ as:
\begin{equation}
\begin{split}
F_{f} &= \frac{1}{2} \left( \frac{1}{2} \sum_{m=1}^{M}\left(\frac{e^{W_{st}^m}}{\sum_{m=1}^{M} e^{W_{st}^m}} \odot F_{f}^m \right) \right. \\
&\quad + \left. \frac{1}{2} \sum_{m=1}^{M}\left(\frac{e^{W_{dy}^m}}{\sum_{m=1}^{M} e^{W_{dy}^m}} \odot F_{f}^m \right) \right) + \frac{1}{2}F_{s},
\end{split}
\end{equation}
where $W_{st}^m, W_{dy}^m \in \mathbb{R}^{C}$ are the $m$-th rows of $W_{st}$ and $W_{dy}$, respectively. The residual connections ensure that the final feature preserves both the original signal and adaptively filtered spectral components. The combined static-dynamic weighting enables robust and adaptive channel-wise integration of multi-scale frequency cues.

\subsection{Adjacent Action Discrepancy Loss}
\label{sec:AADL}

Serving as the ``surgical objective'' for the ``scalpel'', the Adjacent Action Discrepancy Loss (AADL) is designed to maximize the spectral discrepancy between adjacent action segments after filtering, thereby enhancing class discriminability and boundary clarity, as illustrated in Fig.~\ref{fig:4}(b).

\textbf{Amplitude Spectrum Extraction of Action Segments.} 
First, the filtered feature $F_f$ is segmented according to ground-truth action boundaries, resulting in $N$ action segment features: $F_a^1 \in \mathbb{R}^{C \times T_1 \times V}, F_a^2 \in \mathbb{R}^{C \times T_2 \times V}, \dots, F_a^N \in \mathbb{R}^{C \times T_N \times V}$. Each segment $F_a^n$ is then transformed into the frequency domain, its amplitude spectrum is computed, and Linear Interpolation (LI) is applied to unify the frequency resolution to a fixed length $S_f$:
\begin{equation}
F_b^n = \mathrm{LI}(| \mathcal{F}(F_a^n) |),
\end{equation}
where $|\cdot|$ denotes the amplitude of the complex-valued spectrum. Due to the properties of the FFT, regardless of the frame length $T_n$, the frequency axis always spans $[0, f_s)$, where $f_s$ is the sampling rate. The effective physical frequency range is $[0, f_s/2]$, and $T_n$ only affects the frequency resolution $\Delta f = f_s / T_n$. Therefore, interpolated spectra $F_b^1, \dots, F_b^N \in \mathbb{R}^{C \times S_f \times V}$ share a consistent frequency axis, enabling meaningful computation of differences between adjacent actions $F_b^n - F_b^{n-1}$.

\textbf{Loss Computation.} 
AADL computes the discrepancy between adjacent action spectral features by applying absolute difference, averaging, scaling, $\mathrm{tanh}$ nonlinearity, and logarithmic function to encourage spectral separation. The loss $\mathcal{L}_{AAD}$ is formulated as:
\begin{equation}
\mathcal{L}_{AAD}= \frac{1}{N-1} \sum_{n=2}^{N}-\log\bigl(\tanh (\alpha \cdot \mathbb{E}|F_b^n - F_b^{n-1}|)\bigr),
\end{equation}
where $|\cdot|$ denotes element-wise absolute value and $\mathbb{E}(\cdot)$ computes the average across all elements. $\alpha$ is the scaling factor. Therefore, minimizing $\mathcal{L}_{AAD}$ can increase the difference between adjacent actions.

\subsection{Frequency-Aware Channel Mixer}

The Frequency-Aware Channel Mixer (FACM), integrated within the temporal modeling block, enhances channel interactions from a frequency-domain perspective. In conjunction with temporal interactions, it facilitates comprehensive representation modeling, as shown in Fig.~\ref{fig:4}(c).

For the $l$-th block, FACM begins with the temporally modeled feature $F_{t1}^l \in \mathbb{R}^{C \times T}$, which undergoes FFT. The real and imaginary components $R_0^l, I_0^l \in \mathbb{R}^{C \times S}$ are separated and concatenated, followed by two point-wise convolution layers to mix channels:
\begin{equation}
R^l, I^l = \mathrm{Split}(W_{c2} \cdot W_{c1} \cdot \mathrm{Concat}[R_0^l,I_0^l]),
\end{equation}
where $\mathrm{Concat}$ forms $\mathbb{R}^{2\times C \times S}$, $\mathrm{Split}$ reverts to two $\mathbb{R}^{C \times S}$ components, and $W_{c1}, W_{c2} \in \mathbb{R}^{C \times C}$ are point-wise convolutions. The transformed $R^l, I^l  \in \mathbb{R}^{C \times S}$ are recombined into complex spectra and projected back to the time domain via inverse FFT, yielding frequency-aware channel mixed feature $F_{t2}^l \in \mathbb{R}^{C \times T}$.
This convolution strategy is mathematically equivalent to applying a shared linear transformation to the real and imaginary parts, which is analytically equivalent to a direct linear transformation of the complex spectrum, i.e. $W \cdot R_0^l + W \cdot  j I_0^l = W \cdot F^l$. 
This approach maintains complex linear plasticity while avoiding nonlinear decomposition loss (e.g., amplitude/phase). As a result, it enables parameter-efficient combined learning of real and imaginary components while maintaining spectral integrity, thereby improving generalization.

%% file: sec/4_experiment.tex
\section{Experiment}
\label{sec:experiment}

\begin{table*}[tb]
    \footnotesize
    \centering
    \caption{Comparison with the latest results on PKU-MMD v2 (X-view) and PKU-MMD v2 (X-sub).  \textbf{Bold} and \underline{underline} indicate the best and second-best results in each column. FLOPs denote the computational complexity of the model when the input is $X \in \mathbb{R}^{12 \times 6000 \times 25 }$ on the PKU-MMD dataset. }
    \setlength{\tabcolsep}{4pt}
    \renewcommand{\arraystretch}{0.85}
    \begin{tabular}{c|l|cc|ccccc|ccccc}
        \toprule
        & Dataset & & & \multicolumn{5}{c|}{\textbf{PKU-MMD v2 (X-view)}} & \multicolumn{5}{c}{\textbf{PKU-MMD v2 (X-sub)}} \\
        \cmidrule(lr){5-9} \cmidrule(lr){10-14}
        & Metric & FLOPs$\downarrow$ & Param.$\downarrow$ & Acc$\uparrow$ & Edit$\uparrow$ & \multicolumn{3}{c|}{F1@\{10,25,50\}$\uparrow$} & Acc$\uparrow$ & Edit$\uparrow$ & \multicolumn{3}{c}{F1@\{10,25,50\}$\uparrow$} \\
        \midrule
        \multirow{5}{*}{\begin{sideways}\textbf{VTAS}\end{sideways}}
        & MS-TCN~\cite{MS-TCN} & 3.27G & 0.54M & 58.2 & 56.6 & 58.6 & 53.6 & 39.4 & 65.5 & - & - & - & 46.3 \\
        & MS-TCN++~\cite{MS-TCN++} & 5.37G & 0.90M & 59.1 & 59.4 & 61.6 & 56.1 & 42.0 & 66.0 & 66.7 & 69.6 & 65.1 & 51.5  \\
        & ASFormer~\cite{Asformer} & 6.34G & 1.04M & 64.8 & 62.6 & 65.1 & 60.0 & 45.8 & 68.3 & 68.1 & 71.9 & 68.2 & 54.5 \\
        & ASRF~\cite{ASRF} & 7.15G & 1.20M & 60.4 & 59.3 & 62.5 & 58.0 & 46.1 & 67.7 & 67.1 & 72.1 & 68.3 & 56.8 \\
        & LTContext~\cite{LTContext} & 3.71G & 0.56M & 62.3 & 59.6 & 62.9 & 57.8 & 42.6 & 68.8 & 68.2 & 72.1 & 68.5 & 55.4 \\
        
        \midrule
        \multirow{8}{*}{\begin{sideways}\textbf{STAS}\end{sideways}}
        & MS-GCN~\cite{MS-GCN} & 40.2G & 0.65M & 65.3 & 58.1 & 61.3 & 56.7 & 44.1 & 68.5 & - & - & - & 51.6 \\
        & MTST-GCN~\cite{MTST-GCN} & 141.9G & 2.90M & 66.5 & 64.0 & 67.1 & 62.4 & 49.9 & 70.0 & 65.8 & 68.5 & 63.9 & 50.1 \\
        & DeST-{\tiny TCN}~\cite{DeST} & 6.19G & 0.78M & 62.4 & 58.2 & 63.2 & 59.2 & 47.6 & 67.6 & 66.3 & 71.7 & 68.0 & 55.5 \\
        & DeST-{\tiny Former}~\cite{DeST} & 8.17G & 1.11M & 67.3 & 64.7 & 69.3 & 65.6 & 52.0 & 70.3 & 69.3 & 74.5 & 71.0 & 58.7 \\
        & ME-ST~\cite{ME-ST} & 125.93G & 3.42M & \underline{74.1} & \underline{70.5} & \underline{76.6} & \underline{73.2} & \underline{62.4} & 68.5 & 67.2 & 72.3 & 68.8 & 58.1 \\
        & LaSA~\cite{LaSA} & 11.65G & 1.60M & 69.5 & 67.8 & 72.9 & 69.2 & 57.0 & 73.5 & \underline{73.4} & 78.3 & 74.8 & 63.6 \\
        & LPL~\cite{LPL}& 20.77G & 3.23M & 70.0 & 67.7 & 73.3 & 70.0 & 58.5 & \underline{74.7} & 72.7 & \underline{78.6} & \underline{75.8} & \underline{64.3} \\
        
        \cmidrule(lr){2-14}
        & \textbf{Spectral Scalpel} & 11.56G & 1.44M & \textbf{76.5} & \textbf{74.5} & \textbf{80.0} & \textbf{77.3} & \textbf{67.2} & \textbf{75.4} & \textbf{74.5} & \textbf{79.7} & \textbf{76.8} & \textbf{66.6} \\
        \bottomrule
    \end{tabular}
    \label{tab:SOTA1}
\end{table*}

\subsection{Setup}
\label{sec:setup}
\textbf{Datasets and Evaluation Metrics.} We evaluate Spectral Scalpel on five public datasets: PKU-MMD v2 (X-sub and X-view)~\cite{PKU-MMD}, MCFS-130~\cite{MCFS}, LARa~\cite{LARA}, and TCG-15~\cite{tcg}. PKU-MMD v2 is a daily activity dataset comprising 52 action classes, with two standard evaluation protocols: cross-subject (X-sub) and cross-view (X-view). MCFS-130 is a motion-centric figure skating dataset containing 130 fine-grained action categories. LARa is a warehouse activity dataset with 8 categories. TCG-15 is a traffic control gesture dataset consisting of 15 gesture classes. We adopt widely used metrics for TAS, including frame-wise accuracy (Acc), segmental edit distance, and segmental F1 scores at Intersection-over-Union (IoU) thresholds of 10\%, 25\%, and 50\%, denoted as F1@\{10, 25, 50\}.

\textbf{Implementation Details.} 
The feature channel dimension is set to $C = 64$. MASF uses $M = 4$ spectral filters with a maximum filter length of $S_{\text{max}} = 64$. In dynamic weighting, the pooling length is $T_d = 64$ and the projection dimension is $D = 4$. For AADL, the interpolated spectral length is fixed at $S_f = 32$, with a scaling factor $\alpha = 100$. All models are optimized using Adam with a learning rate of 0.001. Following prior works, batch sizes are set to 8 (PKU-MMD, LARa), 2 (TCG-15), and 1 (MCFS-130). Models are trained for 300 epochs, except LARa, which is trained for 60 epochs due to faster convergence. All experiments are performed on a single RTX 3090 GPU.

\subsection{Comparison with State-of-the-Art Methods}
\label{sec:comparison}

\textbf{Quantitative Comparison.}
We compare Spectral Scalpel with recent state-of-the-art VTAS and STAS models on five public datasets, using identical skeleton-based input features to ensure fairness. As shown in Tables~\ref{tab:SOTA1} and~\ref{tab:SOTA2}, Spectral Scalpel consistently achieves state-of-the-art performance across nearly all metrics, while maintaining competitive computational efficiency. For the most representative segment-level metric, F1@50, it surpasses previous best results by notable margins: +4.8\% on PKU-MMD (X-view), +2.3\% on PKU-MMD (X-sub), +1.0\% on MCFS-130, +0.9\% on TCG-15, and +0.8\% on LARa. Furthermore, it achieves lower FLOPs and fewer parameters than previous top-performing models, demonstrating superior efficiency. These results underscore Spectral Scalpel’s strong segmentation capability and high efficiency, validating the effectiveness of its frequency-aware filtering and channel mixing mechanisms.

\begin{table*}[tb]
    \footnotesize
    \centering
    \caption{Comparison with the latest results on MCFS-130, TCG-15, and LARa. \textbf{Bold} and \underline{underline} indicate the best and second-best results in each column.}
    \setlength{\tabcolsep}{4pt}
    \renewcommand{\arraystretch}{0.85}
    \begin{tabular}{c|l|ccccc|ccccc|ccccc}
        \toprule
        & Dataset & \multicolumn{5}{c|}{\textbf{MCFS-130}} & \multicolumn{5}{c|}{\textbf{TCG-15}} & \multicolumn{5}{c}{\textbf{LARa}} \\
        \cmidrule(lr){3-7} \cmidrule(lr){8-12} \cmidrule(lr){13-17}
        & Metric & Acc$\uparrow$ & Edit$\uparrow$ & \multicolumn{3}{c|}{F1@\{10, 25, 50\}$\uparrow$} & Acc$\uparrow$ & Edit$\uparrow$ & \multicolumn{3}{c|}{F1@\{10, 25, 50\}$\uparrow$} & Acc$\uparrow$ & Edit$\uparrow$ & \multicolumn{3}{c}{F1@\{10, 25, 50\}$\uparrow$} \\
        \midrule
        \multirow{5}{*}{\begin{sideways}\textbf{VTAS}\end{sideways}}
        & MS-TCN~\cite{MS-TCN} & 65.7 & 54.5 & 56.4 & 52.2 & 42.5 & 85.5 & 67.7 & 70.0 & 67.3 & 59.1 & 65.8 & - & - & - & 39.6 \\
        & MS-TCN++~\cite{MS-TCN++} & 65.5 & 59.8 & 60.1 & 55.7 & 46.1 & 86.0 & 71.0 & 74.6 & 71.7 & 62.9 & 71.7 & 58.6 & 60.1 & 58.6 & 47.0  \\
        & ASFormer~\cite{Asformer} & 67.5 & 69.1 & 68.3 & 64.0 & 55.1 & 82.3 & 69.1 & 71.3 & 67.7 & 57.6 & 72.2 & 62.2 & 66.1 & 61.9 & 49.2 \\
        & ASRF~\cite{ASRF} & 65.6 & 65.6 & 66.7 & 62.3 & 51.9 & 83.9 & 65.3 & 71.5 & 69.9 & 62.4 & 71.9 & 63.0 & 68.3 & 65.3 & 53.2 \\
        & LTContext~\cite{LTContext} & 72.2 & 74.2 & 74.6 & 70.0 & 59.6 & 86.9 & 75.3 & 79.1 & 75.9 & 67.4 & 71.4 & 62.5 & 68.1 & 64.3 & 52.1 \\
        
        \midrule
        \multirow{8}{*}{\begin{sideways}\textbf{STAS}\end{sideways}}
        & MS-GCN~\cite{MS-GCN} & 64.9 & 52.6 & 52.4 & 48.8 & 39.1 & 86.6 & 68.7 & 70.9 & 69.7 & 59.4 & 65.6 & - & - & - & 43.6 \\
        & IDT-GCN~\cite{IDT-GCN} & 68.6 & 70.2 & 70.7 & 67.3 & 58.6 & - & - & - & - & - & - & - & - & - & - \\
        & MTST-GCN~\cite{MTST-GCN} & - & - & - & - & - & 85.9 & 68.9 & 71.3 & 68.3 & 60.2 & 73.7 & 58.6 & 63.8 & 59.4 & 47.6 \\
        & DeST-{\tiny TCN}~\cite{DeST} & 70.5 & 73.8 & 74.0 & 70.7 & 61.8 & 87.1 & 73.7 & 78.7 & 76.9 & 69.6 & 72.6 & 63.7 & 69.7 & 66.7 & 55.8 \\
        & DeST-{\tiny Former}~\cite{DeST} & 71.4 & 75.8 & 75.8 & 72.2 & 63.0 & 88.1 & 75.7 & 80.0 & 78.2 & 71.2 & 75.1 & 64.2 & 70.3 & 68.0 & 57.7 \\
        & ME-ST~\cite{ME-ST} & - & - & - & - & - & 88.7 & 77.3 & 81.8 & 78.6 & 72.4 & 74.2 & 65.0 & 71.0 & 68.2 & 57.1  \\
        & LaSA~\cite{LaSA} & \underline{72.6} & \textbf{79.3} & \underline{79.3} & \underline{75.8} & \underline{66.6} & 88.7 & 77.4 & 81.2 & 79.5 & 73.3 & 75.3 & \textbf{65.7} & 71.6 & 69.0 & 57.9 \\
        & LPL~\cite{LPL} & - & - & - & - & - & \underline{88.8} & \textbf{77.4} & \underline{81.9} & \underline{80.0} & \underline{73.8} & \underline{76.1} & \underline{65.2} & \textbf{72.3} & \underline{70.0} & \underline{58.6} \\
        
        \cmidrule(lr){2-17}
        & \textbf{Spectral Scalpel} & \textbf{72.7} & \underline{78.0} & \textbf{79.7} & \textbf{76.5} & \textbf{67.6} & \textbf{89.0} & \underline{77.0} & \textbf{81.9} & \textbf{80.0} & \textbf{74.7} & \textbf{76.2} & 64.6 & \underline{72.1} & \textbf{70.1} & \textbf{59.4} \\
        \bottomrule
    \end{tabular}
    \label{tab:SOTA2}
\end{table*}

\begin{figure*}[t]
  \centering
   \includegraphics[width=0.95\linewidth]{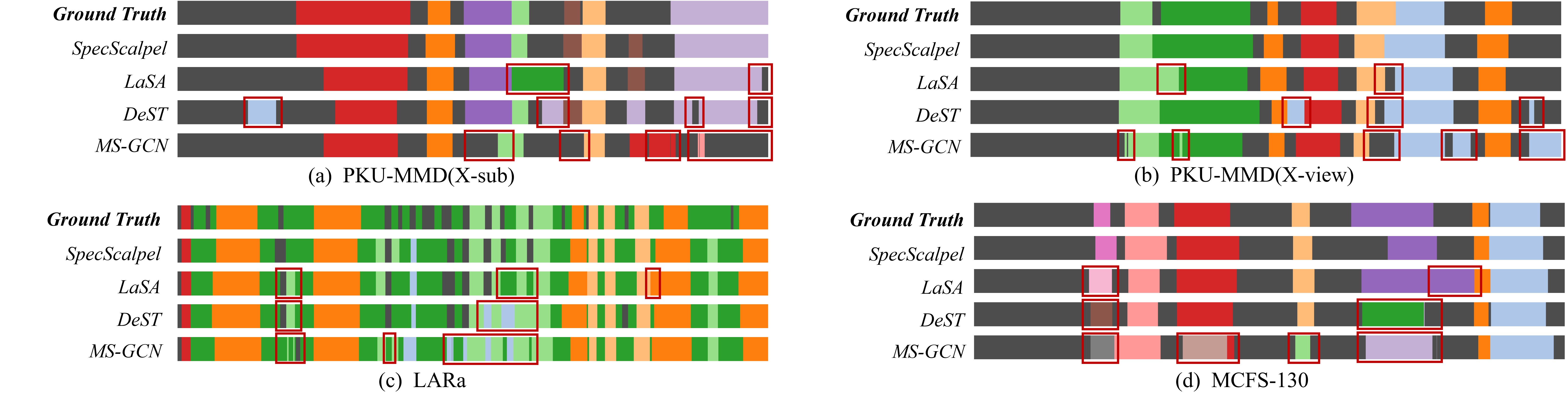}

   \caption{Qualitative results of action segmentation on the PKU-MMD v2 (X-sub, X-view), LARa and MCFS-130 datasets. Different colors represent distinct action classes. Red boxes highlight segmentation errors in other methods compared to Spectral Scalpel.}
   \label{fig:5}
\end{figure*}

\textbf{Qualitative Comparison.}
We present a qualitative comparison of segmentation results between Spectral Scalpel and other state-of-the-art STAS methods in Fig.~\ref{fig:5}. Compared to Spectral Scalpel, existing methods such as LaSA~\cite{LaSA}, DeST~\cite{DeST}, and MS-GCN~\cite{MS-GCN} suffer from more frequent misclassifications, under-segmentation, and boundary shifts, and MS-GCN also exhibits over-segmentation issues. In contrast, while Spectral Scalpel still exhibits minor boundary imprecisions and misclassifications, it produces predictions that are more closely aligned with the ground truth. These observations highlight the superior potential of its discrepancy-guided spectral filtering and channel mixing mechanisms.

\begin{figure}[t]
  \centering
   \includegraphics[width=0.80\linewidth]{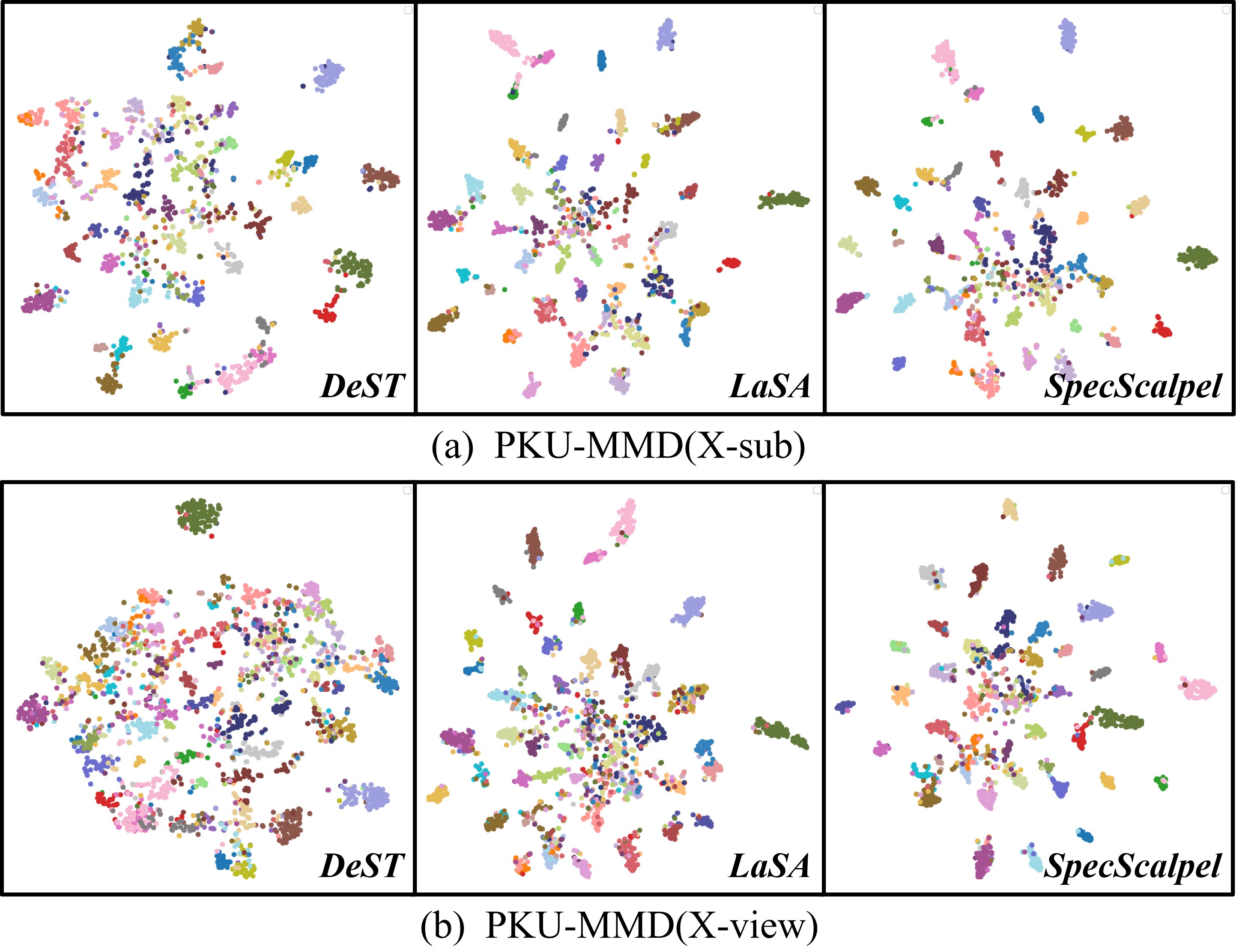}

   \caption{Visualization of the representation space on the PKU-MMD v2 datasets. Each point represents a feature of an action segment, which is colored according to its class label. As observed, compared to others, Spectral Scalpel engenders a more distinctly structured semantic feature space.}
   \label{fig:6}
\end{figure}

\textbf{Visualization of Representation Space.}
We further visualize the representation space of Spectral Scalpel, LaSA~\cite{LaSA}, and DeST~\cite{DeST} on PKU-MMD, as shown in Fig.~\ref{fig:6}. Specifically, we extract all test-set inference representations $F_R$ from the model, segment them according to ground-truth boundaries, and compute the action segment representation $F_{ac} \in \mathbb{R}^C$ by temporal averaging. These vectors are then projected into 2D using t-SNE for visualization. Compared to DeST, LaSA exhibits improved but still entangled distributions, while Spectral Scalpel yields more compact intra-class clusters and more uniformly separated inter-class regions. This clearly demonstrates the effectiveness of discrepancy-guided spectral filtering in enhancing feature discriminability across action segments.

\subsection{Ablation Studies and Analysis}
\label{sec:ablation}

\begin{table}[tb]
    \footnotesize
    \centering
    \setlength{\tabcolsep}{4pt}
    \renewcommand{\arraystretch}{1.0}
    
    \caption{{Ablation study of MASF, AADL, and FACM modules on PKU-MMD (X-sub) dataset.}}
    \begin{tabular}{l|ccccc}
        \toprule
        Components & Acc & Edit & \multicolumn{3}{c}{F1@\{10, 25, 50\}} \\
        \midrule
        Baseline & 73.6& 73.0& 78.2& 74.6& 64.3
\\
        \midrule
        Baseline+MASF & 74.5& 73.7& 78.7& 75.4& 65.6
\\
        Baseline+AADL & 74.8& 73.9& 79.0& 75.5& 65.5
\\
        Baseline+FACM & 74.0& 73.2& 78.0& 74.8& 65.7
\\
        \midrule
        Baseline+MASF+AADL & 75.1& 73.9& 79.3& 76.3& 66.0
\\
        Baseline+MASF+FACM & 74.8& 74.1& 79.5& 76.5& 66.2
\\
        Baseline+AADL+FACM & 74.9& 74.3& 79.6& 76.6& 66.3
\\
        \midrule
        Baseline+All (SpecScalpel) & \textbf{75.4}& \textbf{74.5}& \textbf{79.7}& \textbf{76.8}& \textbf{66.6}\\
        \bottomrule
    \end{tabular}
    \label{tab:Allmodules}
\end{table}

\begin{figure}[tb]
    \centering
    \includegraphics[width=\linewidth]{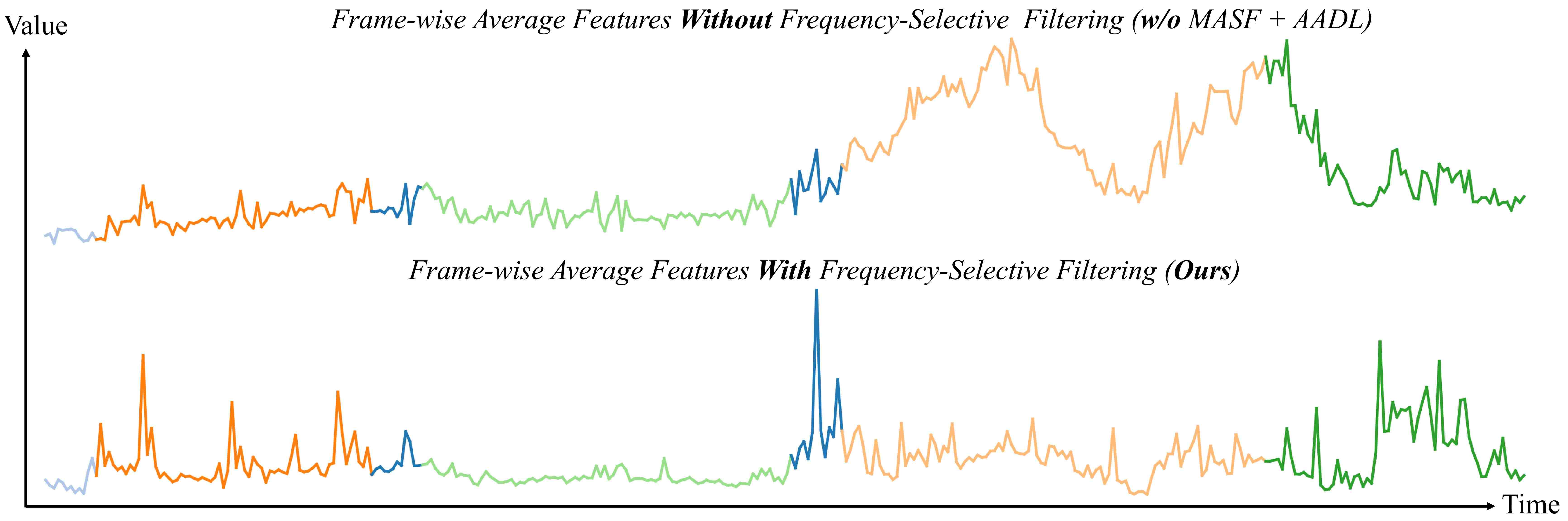}
    
    \caption{Comparison of unfiltered (top) and filtered (bottom) frame-wise average features. X-axis denotes time, y-axis shows per-frame mean value, and colors indicate different actions.}
    \label{fig:7}
\end{figure}

To assess the individual contributions and complementary effects of MASF, AADL, and FACM, we performed an incremental ablation study, as shown in Tab.~\ref{tab:Allmodules}. The Baseline is built upon DeST~\cite{DeST}, equipped with a CTR-GCN~\cite{CTR-GCN}-style adaptive GCN and a LaSA~\cite{LaSA}-inspired Action–Text Contrastive Loss, corresponding to the model configuration shown in Fig.~\ref{fig:3} with MASF, AADL, and FACM removed. Integrating any single module into the baseline yields consistent performance gains, demonstrating the filtering capability of MASF, the discrepancy-guided enhancement of AADL, and the channel evolution role of FACM. Moreover, combining any two modules leads to further improvements, confirming their complementary nature. The full integration of all three modules constitutes our complete model, which achieves the best performance. 

In terms of efficiency, MASF, which performs filtering through a single-stage Hadamard product, adds only +0.01G FLOPs, +0.006M parameters, and increases training time by approximately 1\% compared to the baseline. AADL, being a training-only auxiliary loss, increases training time by around 21\% without introducing any additional FLOPs or parameters (thus leaving inference speed unaffected). FACM, which performs lightweight channel interactions, adds merely +0.49G FLOPs, +0.083M parameters, and about 1\% training time overhead. These results demonstrate that all three modules deliver significant performance improvements with minimal computational overhead.

As illustrated in Fig.~\ref{fig:7}, temporal visualizations of frame-wise feature activations (averaged across spatial and channel dimensions) demonstrate that frequency-selective filtering effectively amplifies inter-action discrepancies. The differences in frequency and amplitude between actions become more pronounced—for instance, the 2nd–5th action segments exhibit highly similar and indistinguishable frequency–amplitude patterns without filtering, which become clearly separated with filtering, especially for the orange (2nd), green (4th), and blue (5th) segments with distinct amplitude variations. Moreover, the filtering process successfully suppresses the shared low-frequency components between the yellow (6th) and green (7th) segments, leading to a clearer contrast in their waveforms and amplitudes.

%% file: sec/5_conclusion.tex
\section{Conclusion}
\label{sec:conclusion}

This paper presents Spectral Scalpel, a discrepancy-guided frequency-selective filtering framework designed to enhance inter-class discriminability and sharpen action boundaries. It introduces a Multi-scale Adaptive Spectral Filter (MASF) and Adjacent Action Discrepancy Loss (AADL) to amplify discrepancies between adjacent actions through spectral filtering, while a Frequency-Aware Channel Mixer (FACM) facilitates frequency-guided channel evolution during temporal modeling. To our knowledge, Spectral Scalpel is the first to incorporate frequency-domain analysis into STAS, achieving state-of-the-art performance with competitive efficiency. 
Despite its strong performance, Spectral Scalpel still exhibits occasional misclassifications and boundary shifts, indicating room for improvement. Future work should focus on adaptive local filtering and multi-level filtering, contrastive learning between positive and negative frequency patterns, and the incorporation of frequency priors to enhance model generalization.

%% file: sec/X_suppl.tex
\clearpage
\setcounter{page}{1}
\maketitlesupplementary

This appendix provides supplementary material for ``Spectral Scalpel: Amplifying Adjacent Action Discrepancy via Frequency-Selective Filtering for Skeleton-Based Action Segmentation.'' We present detailed methodological supplements, comprehensive experimental settings, and further analytical results. Sec.~\ref{app:method} elaborates on architectural components not detailed in the main paper, including spatial modeling, frequency-domain transforms, temporal modeling, prediction refinement, and loss functions. Sec.~\ref{app:setup} outlines the experimental setup, covering datasets, metrics, and implementation details. Sec.~\ref{app:compare} provides additional comparisons, including quantitative clustering analysis, per-class performance breakdowns, inference speed comparison, and a series of robustness evaluations. Sec.~\ref{app:ablation} details extensive ablation studies validating the design of our core components: MASF, AADL, and FACM, along with other architectural choices. Finally, Sec.~\ref{app:discussion} discusses the work's limitations, potential future research directions, and broader societal impacts.

\section{Method Supplement}
\label{app:method}

This section provides a detailed exposition of the architectural components and methodologies employed in Spectral Scalpel. We elaborate on the spatial modeling pipeline (Sec.~\ref{app:method:spatial}), the foundational Fast Fourier Transform (FFT) operations and their spectral properties (Sec.~\ref{app:method:frequency}), the auxiliary temporal modeling blocks (Sec.~\ref{app:method:temporal}), the prediction refinement branches (Sec.~\ref{app:method:prediction}), and the comprehensive formulation of the loss functions (Sec.~\ref{app:method:loss}).

\subsection{Spatial Modeling Supplement}
\label{app:method:spatial}

In the spatial modeling stage, Spectral Scalpel follows the general paradigm introduced by previous works~\cite{TRG-Net}, where the input skeleton sequence $X \in \mathbb{R}^{C_0 \times T \times V}$ is first processed via multi-scale Graph Convolutional Networks (GCNs)~\cite{DeST, MS-G3D} to extract initial spatial representations $F_{s0} \in \mathbb{R}^{C \times T \times V}$. These features are then further refined through temporal-wise and channel-wise dynamic GCNs~\cite{MTST-GCN, TRG-Net, CTR-GCN}, resulting in the final spatial feature $F_s \in \mathbb{R}^{C \times T \times V}$.

\textbf{Multi-Scale GCN for Spatial Topology Encoding.}
We adopt the multi-scale GCN mechanism originally proposed in~\cite{MS-G3D} and introduced into action segmentation by~\cite{DeST} to capture the human skeleton's multi-hop connectivity patterns. To this end, we first define a set of $k$-adjacency matrices $A^k \in \{0,1\}^{V \times V}$, where each matrix encodes joint pairs at exactly $k$-hop distances in the skeletal graph:
\begin{equation}
A^{k}_{ij} = 
\begin{cases} 
1, & \text{if } d(\alpha_i,\alpha_j) = k \text{ or } i = j, \\
0, & \text{otherwise},
\end{cases}
\end{equation}
where $d(\alpha_i, \alpha_j)$ denotes the shortest path length (i.e., number of bones) between joints $\alpha_i$ and $\alpha_j$. For example, the distance between the left wrist and left elbow is $1$, while the distance between the left wrist and left shoulder is $2$.

Next, we construct a unified multi-scale adjacency matrix $A^{MS} \in \{0,1\}^{V \times KV}$ by concatenating all normalized $k$-adjacency matrices up to scale $K$:
\begin{equation}
\begin{split}
A^{MS} = & [(\tilde{D}^1)^{-\frac{1}{2}} A^{1} (\tilde{D}^1)^{-\frac{1}{2}}] \oplus \cdots \\
 & \oplus [(\tilde{D}^K)^{-\frac{1}{2}} A^{K} (\tilde{D}^K)^{-\frac{1}{2}}],
\end{split}
\end{equation}
where $\oplus$ denotes concatenation along the second (column) dimension, and each $\tilde{D}^k$ is a diagonal normalization matrix with elements defined as $\tilde{D}^k_{ii} = \sum_j A^k_{ij} + \epsilon$, where a small constant $\epsilon = 0.001$ is added to prevent division by zero due to empty rows.

Given the input feature $X$, the multi-scale GCN produces spatial features $F_{s0}$ via:
\begin{equation}
F_{s0} = \mathrm{ReLU} \left( \mathrm{reshape} \left[ (A^{MS} + B) \cdot X \right] \cdot W_{s0} \right),
\end{equation}
where $B \in \mathbb{R}^{V \times KV}$ is a learnable matrix that adaptively encodes non-local joint correlations, and $\mathrm{reshape}(\cdot)$ denotes a reshaping operation that transforms the intermediate feature from $\mathbb{R}^{C_0 \times T \times KV}$ to $\mathbb{R}^{KC_0 \times T \times V}$. The operator $W_{s0} \in \mathbb{R}^{1 \times 1 \times KC_0 \times C}$ is a point-wise convolution used to fuse and project the concatenated features to the desired channel dimension.

\textbf{Temporal-wise and Channel-wise Dynamic GCNs for Fine-Grained Spatial Adaptation.}
To further capture the intricate and dynamic inter-joint dependencies in a fine-grained and feature-adaptive manner, Spectral Scalpel integrates both temporal-wise and channel-wise dynamic GCNs into its spatial modeling pipeline. The concept of channel-wise dynamic GCN was originally introduced in CTR-GCN~\cite{CTR-GCN}, enabling independent spatial relation modeling across feature channels. Building on this, subsequent works~\cite{MTST-GCN, TRG-Net} extended the idea to the temporal domain, developing temporal-wise dynamic GCNs to model frame-specific spatial structures.

Formally, given the initial spatial feature $F_{s0} \in \mathbb{R}^{C \times T \times V}$, the model first applies two parallel convolutional heads to produce intermediate embeddings: $P, Q \in \mathbb{R}^{C_1 \times T \times V}$. To capture both temporal- and channel-specific structural cues, two types of pooling operations are performed. Channel pooling along the channel dimension yields: $P^{T}, Q^{T} \in \mathbb{R}^{T \times V}$, while temporal pooling along the temporal axis produces: $P^{C}, Q^{C} \in \mathbb{R}^{C_1 \times V}$.

Next, we compute two adaptive graph structures using cross-joint mean differences to measure pairwise joint dissimilarities in each domain. Specifically, the temporal-wise dynamic graph $G^{T} \in \mathbb{R}^{T \times V \times V}$ and the channel-wise dynamic graph $G^{C} \in \mathbb{R}^{C_1 \times V \times V}$ are computed as:
\begin{equation}
G^{T}_{t,i,j} = P^{T}_{t,i} - Q^{T}_{t,j}, \quad
G^{C}_{c,i,j} = P^{C}_{c,i} - Q^{C}_{c,j},
\end{equation}
where $G^{T}_{t,i,j}$ denotes the relation between joint $i$ and $j$ in the $t$-th frame, and similarly $G^{C}_{c,i,j}$ indicates the relation between joint $i$ and $j$ under the $c$-th channel perspective.

Furthermore, inspired by TRG-Net~\cite{TRG-Net}, we inject semantic structural priors into these dynamic graphs by incorporating a Text-derived Joint Graph (TJG). Specifically, the descriptive texts of the $V$ joints are encoded via a pre-trained BERT~\cite{BERT} model to extract joint embeddings. By calculating the L2 distances among these embeddings and applying inverse normalization, we construct a static semantic graph $G^{txt} \in \mathbb{R}^{V \times V}$. This prior graph $G^{txt}$ is then broadcasted and added element-wise to both $G^{T}$ and $G^{C}$, effectively enriching the data-driven dynamic graphs with explicit anatomical semantics.

The refined input feature $F_{s1} \in \mathbb{R}^{C \times T \times V}$, obtained via a point-wise convolution over $F_{s0}$, is then modulated by these two dynamic graphs through graph convolution along the joint dimension. The final spatially-adaptive feature $F_s \in \mathbb{R}^{C \times T \times V}$  is computed as:
\begin{equation}
F_{s} = \mathrm{ReLU}\left( \mathrm{BN} \left[ F_{s1} \cdot G^{T} + F_{s1} \cdot G^{C} \right] \right) + F_{s0},
\end{equation}
where $\cdot$ denotes batched matrix multiplication over the joint axis $V$, $\mathrm{BN}(\cdot)$ represents batch normalization, and $\mathrm{ReLU}(\cdot)$ is the non-linear activation function. This fusion mechanism enables the model to capture joint dependencies that are adaptively specialized to individual frames and channels, leading to enhanced spatial expressivity and discriminability.

\subsection{Frequency Modeling Supplement}
\label{app:method:frequency}

While frequency modeling has been discussed in the main body, this section provides a detailed exposition of the Fast Fourier Transform (FFT) and its inverse (iFFT), which serve as the fundamental tools for transforming features from the time domain to the frequency domain. We also elaborate on the spectral properties that arise from this transformation, including Hermitian symmetry and the implications of the Nyquist-Shannon Sampling Theorem.

\textbf{Fast Fourier Transform and Spectral Properties.}
To enable frequency-domain modeling of temporal dynamics, we apply the FFT and its inverse, denoted by $\mathcal{F}(\cdot)$ and $\mathcal{F}^{-1}(\cdot)$, respectively. The FFT is performed along the temporal dimension $T$ of the input feature tensor $F \in \mathbb{R}^{C \times T \times V}$ or $F \in \mathbb{R}^{C \times T}$, converting time-domain signals into the frequency domain. The resulting frequency-domain representation lies in the complex space, with shape $\mathbb{C}^{C \times S \times V}$ or $\mathbb{C}^{C \times S}$.

Formally, for a feature of dimension \(\mathbb{R}^{C \times T \times V}\), the FFT is applied to each channel \(c\) and spatial joint \(v\). Its computation aligns with the Discrete Fourier Transform (DFT) and is defined as follows:  
\begin{equation}
\begin{gathered}
\hat{F}_{c,s,v}  = \mathcal{F}(F)_{c,s,v}
= \sum_{t=0}^{T-1} F_{c,t,v} \cdot e^{-2\pi i \cdot \frac{st}{T}},  \\
\quad s = 0, 1, \dots, S-1,
\end{gathered}
\end{equation}
where \(S = T\) (i.e., the output frequency spectrum has the same length as the input signal). To reconstruct the time-domain signal from its frequency-domain representation $\hat{F} \in \mathbb{C}^{C \times S \times V}$, we apply the inverse FFT:
\begin{equation}
\begin{gathered}
F_{c,t,v} = \mathcal{F}^{-1}(\hat{F})_{c,t,v} = \frac{1}{T} \sum_{s=0}^{S-1} \hat{F}_{c,s,v} \cdot e^{2\pi i \cdot \frac{st}{T}}, \\
\quad t = 0, 1, \dots, T-1.
\end{gathered}
\end{equation}
These operations are implemented using the FFT algorithm with a computational complexity of $\mathcal{O}(T \log T)$, which is significantly more efficient than the naive DFT with complexity $\mathcal{O}(T^2)$.

For real-valued input signals, the frequency spectrum exhibits \textit{Hermitian symmetry}, meaning that:
\begin{equation}
\hat{F}_{c,T-s,v} = \overline{\hat{F}_{c,s,v}}, \quad \text{for } s = 1, \dots, T/2,
\end{equation}
where $\overline{(\cdot)}$ denotes complex conjugation. 
This symmetry implies redundancy in the second half of the spectrum, allowing the signal to be fully represented by only the first \(T/2 + 1\) frequency components. Thus, for storage and computational efficiency, the frequency-domain feature \(\hat{F} \in \mathbb{C}^{C \times S \times V}\) can be truncated to \(S = T/2 + 1\) (down from \(S = T\)).  The same FFT/iFFT strategy is applied to features of dimension \(\mathbb{R}^{C \times T}\) for each channel \(c\).

According to the Nyquist-Shannon Sampling Theorem, a signal with maximum frequency $f_{\text{max}}$ must be sampled at a rate $f_s \geq 2f_{\text{max}}$ to avoid aliasing. In our case, the temporal signals are sampled uniformly, and the maximum resolvable frequency---the \textit{Nyquist frequency}---is given by: $f_{\text{Nyquist}} = f_s/2$. The frequency resolution $\Delta$ of the spectrum is determined by the number of time samples \(T\), such that $\Delta = f_s/T$. Consequently, the \(i\)-th frequency bin corresponds to \(f_i = i \cdot f_s/T\).

\subsection{Temporal Modeling Supplement}
\label{app:method:temporal}

In each temporal block of Spectral Scalpel, temporal modeling is comprehensively addressed through a combination of frequency-domain channel evolution and time-domain temporal interaction operations. In addition to the proposed Frequency-Aware Channel Mixer (FACM) for frequency-level channel evolution, each temporal block is equipped with a Linear Transformer for temporal interaction and an adaptive feature fusion mechanism---both structurally similar to prior STAS works~\cite{DeST, TRG-Net}.

Specifically, for the $l$-th temporal block, the Linear Transformer first takes the output feature from the previous block, $F_{t}^{l-1} \in \mathbb{R}^{C \times T}$, and models temporal interactions to yield $F_{t1}^{l} \in \mathbb{R}^{C \times T}$. This is then processed by the proposed FACM to produce $F_{t2}^{l} \in \mathbb{R}^{C \times T}$. Finally, an adaptive feature fusion module integrates $F_{t2}^{l}$ with early-stage temporal features $F_{t0}^{l} \in \mathbb{R}^{C \times T}$, extracted via a separate spatial-channel fusion head, to generate the final output of the block $F_{t}^{l} \in \mathbb{R}^{C \times T}$.

\textbf{Linear Transformer for Temporal Interaction.}
A key challenge in temporal modeling for action segmentation is capturing inter-frame dependencies. Traditional methods based on dilated Temporal Convolutional Networks (TCNs) expand the receptive field progressively, but remain limited by fixed local temporal scopes. In contrast, Transformer-based attention mechanisms offer global context modeling. However, the quadratic complexity of conventional attention with respect to sequence length makes it computationally impractical for dense temporal sequences, especially in action segmentation~\cite{Asformer, UVAST}.

To this end, we adopt the Linear Transformer~\cite{Linformer1, Linformer2}, which reduces the attention complexity from $\mathcal{O}(n^2)$ to $\mathcal{O}(n)$, enabling efficient global temporal modeling. This approach has also been utilized in prior STAS works~\cite{DeST, LaSA, TRG-Net}. Formally, the Linear Transformer layer is defined as:
\begin{equation}
F_{t1}^{l} = \mathrm{ReLU}\left[\phi(Q_t^l) \left(\phi(K_t^l)^\top V_t^l\right) \cdot W_t^l + F_{t}^{l-1} \right],
\end{equation}
where the input $F_{t}^{l-1} \in \mathbb{R}^{C \times T}$ is linearly projected into query, key, and value tensors:
\begin{equation}
Q_t^l = W_{Qt}^l F_{t}^{l-1}, \quad K_t^l = W_{Kt}^l F_{t}^{l-1}, \quad V_t^l = W_{Vt}^l F_{t}^{l-1},
\end{equation}
with $W_{Qt}^l, W_{Kt}^l, W_{Vt}^l \in \mathbb{R}^{C \times C_2}$. The output is projected back using $W_t^l \in \mathbb{R}^{C_2 \times C}$. $\phi(\cdot)$ denotes the sigmoid activation function.

\textbf{Adaptive Feature Fusion for Core Feature Retention.}
To align the frequency-modeled output feature $F_f \in \mathbb{R}^{C \times T \times V}$ with the temporal modeling dimension $\mathbb{R}^{C \times T}$, a spatial-channel fusion operation is employed. Inspired by~\cite{TRG-Net}, a point-wise convolution is first used to reduce the channel dimension, producing an intermediate representation of shape $\mathbb{R}^{C_3\times T \times V}$. This tensor is then reshaped to $\mathbb{R}^{VC_3 \times T}$ by merging the channel and spatial dimensions, followed by another point-wise convolution to obtain $F_{t0}^{l} \in \mathbb{R}^{C \times T}$.
This design allows each channel to be adaptively optimized with respect to the spatial dimension $V$, thereby preserving essential spatial semantics. However, due to the compression from $VC$ to $C$, certain spatial information may be lost. To mitigate this, we employ adaptive feature fusion: we use convolution heads with different weights to generate $l$ features from different perspectives, and fuse the feature $F_{t0}^{l}$ of each perspective with the output $F_{t2}^{l}$ of the FACM in each block. This fusion enriches the feature representation by leveraging multiple perspectives from the original $VC$-dimensional space.
The final fused feature output of the $l$-th block is given by:
\begin{equation}
F_{t}^{l} = \mathrm{GeLU}\left[\left(F_{t0}^{l} \oplus F_{t2}^{l}\right) \cdot W_f \cdot W_l\right] + F_{t2}^{l},
\end{equation}
where $\oplus$ denotes channel-wise concatenation, $W_f \in \mathbb{R}^{1 \times 1 \times 2C \times C}$, $W_l \in \mathbb{R}^{C \times C}$ are both point-wise convolution operators. The additive residual connection with $F_{t2}^{l}$ further enhances gradient flow and representation stability.

\subsection{Prediction Refinement Supplement}
\label{app:method:prediction}

Spectral Scalpel follows the prediction refinement paradigm established by previous STAS works, comprising two complementary branches: class prediction and boundary regression. The class refinement strategy, originally proposed in~\cite{MS-TCN}, has been widely adopted across most action segmentation frameworks. Subsequently,~\cite{ASRF} introduced the boundary regression refinement, which effectively mitigates over-segmentation and improves overall segmentation accuracy.

\textbf{Class Prediction Branch.}
Given the final representation of temporal modeling $F_R$, an initial frame-wise class prediction $Y^c_0 \in \mathbb{R}^{Q \times T}$ is produced by a classification head, where $Q$ denotes the number of classes. This prediction is progressively refined through $S^c$ refinement stages. At the $h$-th stage, the previous prediction $Y^c_{h-1} \in \mathbb{R}^{Q \times T}$ is first projected to a feature $\mathbb{R}^{C \times T}$ via a point-wise convolution. A stack of 10 Linear Transformer layers is then applied to generate a more precise prediction $Y^c_h \in \mathbb{R}^{Q \times T}$. Each Linear Transformer layer adopts a cross-attention mechanism. Specifically, the query and key ($Q_t, K_t$) are computed from the current stage's own features, while the value ($V_t$) is derived from the final representation $\mathbb{R}^{C \times T}$ of the previous stage. The final stage yields the final refined class prediction $Y^c_F \in \mathbb{R}^{Q \times T}$.

\textbf{Boundary Regression Branch.}
Similarly, the regression head produces an initial boundary confidence map $Y^b_0 \in \mathbb{R}^{1 \times T}$ from $F_R$, which is refined across $S^b$ stages. At each $h$-th stage, the previous output $Y^b_{h-1} \in \mathbb{R}^{1 \times T}$ is transformed into $\mathbb{R}^{C \times T}$ via a point-wise convolution, followed by a stack of 10 dilated TCNs to yield the refined boundary prediction $Y^b_h \in \mathbb{R}^{1 \times T}$. The final output from the last stage is denoted as $Y^b_F \in \mathbb{R}^{1 \times T}$.

\subsection{Loss Function}
\label{app:method:loss}

As illustrated in Fig.3, the overall loss function is composed of four main components: (i) the Adjacent Action Discrepancy Loss proposed in this work for supervising the filtered feature representation (detailed in the main paper); (ii) the Action-Text Contrastive Loss adapted from previous language-assisted work~\cite{LaSA}; and two commonly employed losses for action segmentation tasks: (iii) the Boundary Regression Loss, and (iv) the Action Segmentation Loss comprising classification and smoothing components. The detailed formulations of each component are provided below.

\textbf{Action-Text Contrastive Loss.}
Following~\cite{LaSA}, we adopt a contrastive learning strategy to align visual features and textual embeddings. Given the final representation after temporal modeling $F_R \in \mathbb{R}^{C \times T}$, it is first projected into a latent space via a learnable mapping head $\mathbb{R}^{C \times C_t}$. The representation sequence is then segmented into $N$ action segments based on the ground-truth boundaries, and the temporal mean is computed within each segment to yield a set of action-level visual representations $A^F \in \mathbb{R}^{N \times C_t}$.

Simultaneously, we obtain corresponding textual embeddings $A^E \in \mathbb{R}^{N \times C_t}$ for the $N$ action classes using BERT~\cite{BERT} pretrained text encoder applied to the class-wise action descriptions. We compute a pairwise cosine similarity matrix between all visual features and textual embeddings:
\begin{equation}
S^A = 
\begin{bmatrix} 
    sim(A^F_1,A^E_1)  & \cdots & sim(A^F_1,A^E_N)\\
    \vdots & \ddots & \vdots\\
    sim(A^F_N,A^E_1)  & \cdots & sim(A^F_N,A^E_N)
\end{bmatrix},
\end{equation}
where cosine similarity is defined as $\mathrm{sim}(z_i, z_j) = (z_i \cdot z_j) / \|z_i\| \|z_j\|$. To enforce bidirectional alignment, we apply softmax normalization along both rows and columns, resulting in $S^A_f$ (row-normalized) and $S^A_e$ (column-normalized), respectively. A ground-truth similarity matrix $S^{GT} \in \mathbb{R}^{N \times N}$ is constructed, where each entry is 1 for positive (same-class) pairs and 0 otherwise. The loss minimizes the Kullback-Leibler (KL) divergence between the predicted and ground-truth similarity distributions:
\begin{equation}
\mathcal{D}_{KL}(U\| W) = \frac{1}{N^2} \sum_{i=1}^{N} \sum_{j=1}^{N} U_{ij} \log \left( \frac{U_{ij}}{W_{ij}} \right),
\end{equation}
where $U, W \in \mathbb{R}^{N \times N}$. The final action-text contrastive loss is defined as:
\begin{equation}
\mathcal{L}_{AT} = \frac{1}{2}[\mathcal{D}_{KL}(S^{GT}\| S^A_f)+\mathcal{D}_{KL}(S^{GT}\| S^A_e)].
\end{equation}

\textbf{Action Segmentation Loss.}
The action segmentation loss supervises the class prediction branches at all stages. For each stage, the segmentation loss $\mathcal{L}_{as}$ comprises two terms: the standard frame-wise classification cross-entropy loss $\mathcal{L}_{ce}$, and the Gaussian Similarity-weighted Truncated Mean Squared Error (GS-TMSE) loss $\mathcal{L}_{gs\text{-}tmse}$, designed to penalize abrupt changes in predictions while preserving sharp action boundaries:
\begin{equation}
\begin{split}
\mathcal{L}_{as}= \mathcal{L}_{ce} + \mathcal{L}_{gs-tmse} 
= -\frac{1}{T} \sum_{t}{\log(\hat{y}_{t, \hat{c}})} \\
+ \frac{1}{TC} \sum_{t,c} e^{-\frac{\|x_{t}-x_{t-1}\|^2}{2\sigma^2}} \min\left( \left|\log \frac{\hat{y}_{t,c}}{\hat{y}_{t-1,c}}\right|^2, \tau^2 \right),
\end{split}
\end{equation}
where $\hat{y}_{t} \in \mathbb{R}^C$ denotes the predicted class probabilities at frame $t$, $\hat{y}_{t, \hat{c}}$ is the predicted probability for the ground truth class $\hat{c}$, and $x_t$ represents the input feature at frame $t$. By weighting the temporal penalty with the Gaussian similarity of adjacent features, this term adaptively enforces smoothness within uniform action segments while relaxing the penalty at action boundaries to avoid over-smoothing. The standard deviation $\sigma$, which controls the similarity sensitivity, is set to 1.0. Furthermore, a predefined threshold $\tau = 4$ is applied to truncate the squared log-difference, preventing excessively large gradients.

\textbf{Boundary Regression Loss.}
The boundary regression loss supervises the prediction of frame-level action boundaries. At each stage, the loss is computed using binary cross-entropy:
\begin{equation}
\mathcal{L}_{br} = -\frac{1}{T} \sum_{t} \left( b_t \log(\hat{b}_t) + (1 - b_t) \log(1 - \hat{b}_t) \right),
\end{equation}
where $b_t \in \{0,1\}$ is the binary ground truth boundary label at frame $t$, and $\hat{b}_t \in [0,1]$ is the predicted boundary probability.

\textbf{Total Loss.}
The total training objective integrates all the aforementioned losses. The overall loss function is as follows:
\begin{equation}
\mathcal{L} = \sum_{h=0}^{S^c} \mathcal{L}_{as} + \lambda_1 \sum_{h=0}^{S^b} \mathcal{L}_{br} + \lambda_2 \mathcal{L}_{AT} + \lambda_3 \mathcal{L}_{AAD}
\end{equation}
where $\mathcal{L}_{AAD}$ denotes the Adjacent Action Discrepancy loss (defined in the main text), and $\lambda_1$, $\lambda_2$, and $\lambda_3$ are the balancing coefficients, empirically set to 1.0, 0.8, and 1.0, respectively.

\section{Experimental Setup Details}
\label{app:setup}

This section outlines the comprehensive experimental protocol used to validate our model. We describe the datasets and preprocessing procedures (Sec.~\ref{app:setup:datasets}), the standard evaluation metrics (Sec.~\ref{app:setup:evaluation}), and additional implementation-specific details and hyperparameter settings (Sec.~\ref{app:setup:implementation}) to ensure reproducibility.

\subsection{Dataset and Preprocessing}
\label{app:setup:datasets}

We utilize the following datasets and follow the standard preprocessing protocols adopted by previous STAS studies to ensure fair comparison and reproducibility.

\textbf{PKU-MMD v2}~\cite{PKU-MMD} is a large-scale multi-modal action detection dataset recorded at 50 Hz using Kinect V2 sensors. It provides 3-axis spatial coordinates for 25 body joints, covering 1009 action instances across 52 categories, totaling over 50 hours of motion data. We adopt two official partition protocols: (i) Cross-subject (X-sub), where 775 videos are used for training and 234 unseen subjects for testing; and (ii) Cross-view (X-view), where training samples are taken from the middle and right camera views, while testing samples are drawn from the left view. During preprocessing, we compute 6-channel features composed of joint-wise relative positions and temporal displacements based on the 3-axis coordinates. The PKU-MMD v2 dataset is licensed under the Apache 2.0 License.

\textbf{LARa}~\cite{LARA} focuses on capturing typical warehouse manipulation and locomotion behaviors, recorded at 200 Hz using an optical marker-based motion capture system. It provides 3-axis position and orientation data for 19 joints across 377 video samples, categorized into 8 action classes, amounting to 758 minutes of data. In preprocessing, we downsample all sequences to 50 Hz to ensure temporal consistency across datasets. We then extract 12-channel features by computing joint-wise relative positions and orientations, along with their respective temporal displacements. The LARa dataset is licensed under the Creative Commons Attribution Non-Commercial 4.0 International License.

\textbf{MCFS-130}~\cite{MCFS} is a figure skating action dataset designed for sports analysis. It is collected at 30 Hz using the OpenPose toolbox and includes 2D joint coordinates for 25 body joints. The dataset contains 271 video clips covering 130 fine-grained action categories, totaling 17.3 hours. For this dataset, we directly use the 2-axis positions to construct 2-channel relative position features for each joint. The MCFS-130 dataset does not specify a license in its original publication, but our use is strictly non-commercial and properly cited the original work.

\textbf{TCG-15}~\cite{tcg} is a traffic control gesture dataset designed for autonomous driving applications. It is collected at 100 Hz using IMU sensors and records 3-axis positions for 17 joints. The dataset contains 550 annotated action segments across 4 categories, with a total duration of approximately 140 minutes. During preprocessing, we compute 6-channel features by deriving relative joint positions and their corresponding displacements from the raw 3-axis positional data. The TCG-15 dataset is licensed under the MIT License.

\subsection{Evaluation Metrics}
\label{app:setup:evaluation}

In accordance with standard practice in prior work, we evaluate model performance using three complementary metrics: Frame-wise Accuracy (Acc), Segmental Edit Score (Edit), and Segmental F1 Score at intersection-over-union (IoU) thresholds of 0.10, 0.25, and 0.50, denoted as F1@\{10, 25, 50\}. These metrics jointly assess both frame-level precision and segment-level structural correctness.

\textbf{Frame-wise Accuracy (Acc)} computes the proportion of frames whose predicted labels match the ground truth. Formally, it is defined as: $Acc = \frac{1}{T} \sum_{t=1}^{T} \mathbb{I}(y_t = \hat{y}_t)$, where $T$ is the total number of frames, $y_t$ and $\hat{y}_t$ denote the ground-truth and predicted labels at frame $t$, respectively, and $\mathbb{I}(\cdot)$ is the indicator function. Although simple and widely adopted, this metric is overly sensitive to minor misclassifications in long segments and fails to penalize over-segmentation.

\textbf{Segmental Edit Score} evaluates the sequence-level similarity between predicted and ground-truth segmentations by computing the minimum number of edit operations---insertions, deletions, and substitutions---required to convert the predicted action sequence into the ground truth. It is particularly effective in capturing structural discrepancies such as over-segmentation, while remaining agnostic to the duration of individual segments.

\textbf{Segmental F1 Score} assesses segment-level alignment by matching predicted and ground-truth segments using a temporal IoU criterion. A predicted segment is considered a true positive if it shares the same action label and exceeds a specified IoU threshold with any ground-truth segment. Unmatched predictions and ground-truth segments are treated as false positives and false negatives, respectively. Precision and recall are then computed, and the F1 score is defined as: \(F1 = 2 \times \frac{\text{Precision} \times \text{Recall}}{\text{Precision} + \text{Recall}}\).
By incorporating multiple IoU thresholds (i.e., F1@10, F1@25, F1@50), this metric provides a nuanced and robust evaluation of both coarse and fine temporal alignment, while explicitly penalizing fragmented or imprecise segmentations.

\subsection{Additional Implementation Details}
\label{app:setup:implementation}

\begin{table*}[tb]
    \footnotesize
    \centering
    \caption{Comparison of Mean Clustering Metrics (SC, CH, DB) on the PKU-MMD v2 (X-sub, X-view), LARa and MCFS-130 datasets. SC denotes the Silhouette Coefficient Index, CH is the Calinski–Harabasz Index, and DB indicates the Davies–Bouldin Index.}
    \renewcommand{\arraystretch}{1.0}
    \begin{tabular}{c|ccc|ccc|ccc|ccc}
        \toprule
         Dataset & \multicolumn{3}{c|}{PKU-MMD v2 (X-sub)}  & \multicolumn{3}{c|}{PKU-MMD v2 (X-view)} & \multicolumn{3}{c|}{MCFS-130} & \multicolumn{3}{c}{LARa}  \\
        \cmidrule(lr){2-4} \cmidrule(lr){5-7} \cmidrule(lr){8-10} \cmidrule(lr){11-13}
       Metric & SC$\uparrow$ & CH$\uparrow$ & DB$\downarrow$ & SC$\uparrow$ & CH$\uparrow$ & DB$\downarrow$ & SC$\uparrow$ & CH$\uparrow$ & DB$\downarrow$ & SC$\uparrow$ & CH$\uparrow$ & DB$\downarrow$\\
        \midrule 
        DeST~\cite{DeST} & 0.19 & 63.5& 1.59 & 0.08 & 62.6 & 2.07 & 0.01 & 36.6& 2.22 & 0.14 & 722.4& 1.74 
\\
        LaSA~\cite{LaSA} & 0.25 & 79.7& 1.31 & 0.16 & 78.3 & 1.62 & 0.05 & 42.9& 2.07 & 0.24 & \textbf{1245.1}& 1.18 
\\
        \textbf{SpecScalpel (Ours)} & \textbf{0.30} & \textbf{88.6}& \textbf{1.24} & \textbf{0.29} & \textbf{128.9} & \textbf{1.26} & \textbf{0.07} & \textbf{49.3}& \textbf{1.94} & \textbf{0.25} & \underline{1189.7}& \textbf{1.12} 
\\
        \bottomrule
    \end{tabular}
    \label{tab:cluster}
\end{table*}

In addition to the implementation details provided in Sec.~4.1, we present further implementation specifications for the supplementary methods described in Sec.~\ref{app:method}. Notably, all the implementation configurations are fully aligned with those used in the previous STAS framework.

For the multi-scale GCN module, the scale $K$ is set to 13. In the temporal-wise and channel-wise dynamic GCNs, the channel dimension $C_1$ is set to 16. For the FFT operations, we adopt the backward normalization mode, which applies no scaling during transformation.
In the Linear Transformer component, the channel dimensions for the query, key, and value projections are all set to $C_2 = 16$. For the adaptive feature fusion module, the compression operation reduces the feature dimensionality to $C_3 = 8$.
Regarding the prediction refinement, we use $S^c = 1$ stage for the class prediction branch and $S^b = 2$ stages for the boundary regression branch.
For the action-text contrastive loss, text embeddings are derived from natural language descriptions of each action category in the dataset. Specifically, sentence-level embeddings are obtained by averaging the word token embeddings produced by a pretrained BERT encoder~\cite{BERT}, resulting in a fixed embedding dimensionality of $\mathbb{R}^{768}$. To maintain modality alignment, the dimensionality of the action-level visual representation is set to $C_t = 768$, matching the output of the pretrained text encoder. The BERT model is licensed under the Apache 2.0 License.

\section{More Comparisons}
\label{app:compare}

This section presents further comparative analyses to provide a comprehensive quantitative and qualitative understanding of the Spectral Scalpel framework. We first evaluate the discriminability of the learned features via established clustering metrics (Sec.~\ref{app:compare:clustering}) and conduct a detailed per-class analysis to highlight the specific advantages and inherent limitations of our frequency-domain modeling (Sec.~\ref{app:compare:per-class}). Next, we validate the practical efficiency of our approach through an inference speed comparison (Sec.~\ref{app:compare:Speed}). Finally, we present a thorough assessment of the model's robustness against a spectrum of challenging conditions, including spatial noise and joint occlusions (Sec.~\ref{app:compare:robustness}), boundary annotation ambiguity (Sec.~\ref{app:compare:robustnessBA}), and temporal scale variations (Sec.~\ref{app:compare:robustnessTR}).

\subsection{Clustering Performance Comparison}
\label{app:compare:clustering}

\begin{figure*}[tb]
    \centering
    \includegraphics[width=\linewidth]{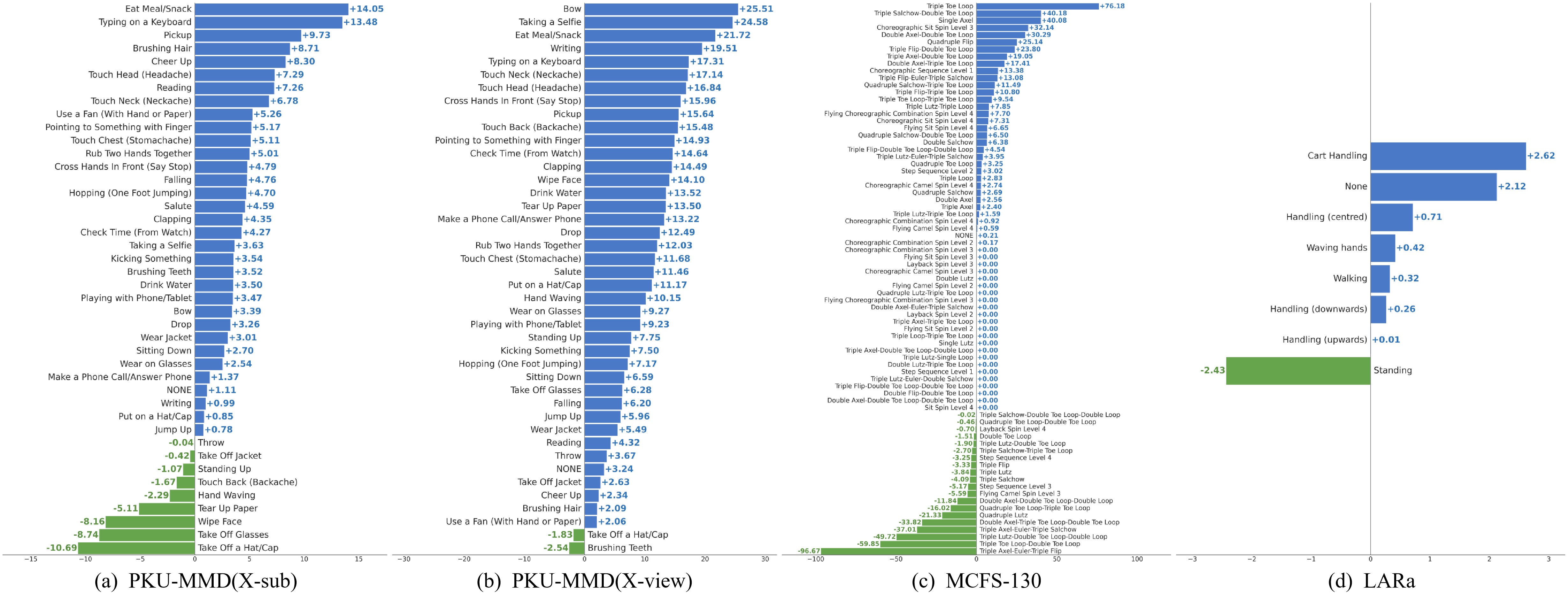}
    
    \caption{Per-class frame-wise F1 score differences between our SpecScalpel and the state-of-the-art LaSA method on the PKU-MMD v2 (X-sub and X-view), LARa, and MCFS-130 datasets. Blue bars represent action classes where SpecScalpel achieves higher F1 scores, whereas green bars correspond to minor declines. The results indicate that SpecScalpel consistently enhances performance across the majority of action categories, with only limited exceptions.}
    \label{fig:8}
\end{figure*}

The t-SNE visualization in Fig.~6 provides only a qualitative assessment of the clustering behavior in the representation space. To more intuitively and quantitatively evaluate the discriminability of the learned features, we compute a series of well-established clustering validity metrics on all high-dimensional action-segment representations $F_{ac} \in \mathbb{R}^C$, including the Silhouette Coefficient (SC) Index, Calinski–Harabasz (CH) Index, and Davies–Bouldin (DB) Index. These metrics jointly measure inter-cluster separability and intra-cluster compactness. Specifically, the Silhouette Coefficient evaluates how well each sample fits within its assigned cluster relative to other clusters (ranging from –1 to 1, with higher values indicating denser and better-separated clusters). The Calinski–Harabasz Index (Variance Ratio Criterion) measures the ratio of between-cluster to within-cluster dispersion (ranging from 0 to $\infty$, with higher values denoting superior clustering). The Davies–Bouldin Index assesses the average similarity between each cluster and its most similar counterpart (ranging from 0 to $\infty$, with lower values indicating better separability).

We compare DeST~\cite{DeST}, LaSA~\cite{LaSA}, and our SpecScalpel across the PKU-MMD v2 (X-sub and X-view)~\cite{PKU-MMD}, MCFS-130~\cite{MCFS}, and LARa~\cite{LARA} datasets, as summarized in Tab.~\ref{tab:cluster}. Our model achieves the best scores on nearly all metrics, clearly demonstrating the strong clustering structure and high discriminability of the learned representations.

\subsection{Analysis of Per-Class Performance}
\label{app:compare:per-class}

To further illustrate the detailed impact of the proposed discrepancy-guided frequency-selective filtering, we compare the per-class frame-wise F1 scores between our SpecScalpel and LaSA~\cite{LaSA}. Notably, both methods are built upon DeST~\cite{DeST} and adopt the Action–Text Contrastive Loss; hence, the primary distinction lies in the frequency-domain modeling introduced by SpecScalpel. This comparison enables a clearer understanding of the types of actions and scenarios in which the proposed frequency-based modeling demonstrates its advantages, as well as those where it faces inherent challenges and limitations.

As shown in Fig.~\ref{fig:8}, SpecScalpel exhibits substantial improvements for action classes characterized by pronounced high-frequency components or sharp boundary transitions, highlighting the effectiveness of discrepancy-guided frequency-selective filtering. Notable gains are observed for high-frequency repetitive actions such as Eat Meal/Snack (+14.05\% on PKU-MMD X-sub / +21.72\% on PKU-MMD X-view), Typing on a Keyboard (+13.48\% on PKU-MMD X-sub / +17.31\% on PKU-MMD X-view), and Brushing Hair (+8.71\% on PKU-MMD X-sub). These actions typically involve rapid, periodic, or quasi-periodic limb motions, producing distinctive high-frequency spectral signatures. The MASF module effectively captures and emphasizes these components, leveraging them as strong discriminative cues.

Similarly, SpecScalpel achieves significant improvements on short-duration actions with abrupt motion boundaries, such as Bow (+25.51\% on PKU-MMD X-view) and Taking a Selfie (+24.58\% on PKU-MMD X-view). These actions exhibit transient, high-frequency spectral responses at the onset and termination phases. The AADL module, by optimizing inter-action spectral discrepancies, enhances such boundary sharpness, thereby outperforming LaSA, which relies solely on temporal-domain modeling.

However, Fig.~\ref{fig:8} also reveals certain failure cases and inherent limitations. SpecScalpel performs less effectively on low-frequency or quasi-static actions, such as Standing (–2.43\% on LARa) and Standing Up (–1.07\% on PKU-MMD X-sub), whose motion characteristics are dominated by static postures and minimal spectral variation concentrated in low and zero-frequency bands. The frequency-selective mechanism may inherently bias toward dynamic  information, leading to insufficient contextual modeling for static actions.

More notably, actions with spectrally similar motion patterns pose additional challenges. For instance, temporally opposite yet spectrally identical actions---such as Take Off a Hat/Cap (–10.69\% on PKU-MMD X-sub / –1.83\%  on PKU-MMD X-view) and Take Off Glasses (–8.16\% on PKU-MMD X-sub)---exhibit nearly indistinguishable dominant frequencies from their corresponding “Put on” counterparts, rendering the discrepancy-guided filtering less discriminative. Furthermore, frequency-degenerate action pairs or compositional actions, such as Quadruple Lutz (–21.33\% on MCFS-130) vs. Triple Lutz (–3.84\% on MCFS-130), or Triple Toe Loop–Double Toe Loop (–59.85\% on MCFS-130) vs. Double Toe Loop (-4.48\% on MCFS-130), differ mainly in repetition count or combination structure, which are inherently difficult to distinguish in the frequency domain. 

In summary, SpecScalpel significantly enhances the representation of high-frequency dynamic motions and sharp-boundary transitions through frequency-domain modeling. Nonetheless, it faces challenges in low-frequency static actions and spectrally indistinguishable motion pairs, revealing its current limitations and providing valuable insights for future work on more adaptive frequency modeling.

\subsection{Inference Speed Comparison}
\label{app:compare:Speed}

To evaluate the practical efficiency of our proposed framework, we measure the average inference speed on the test set using a single NVIDIA RTX 3090 GPU with a batch size of 1. During inference, our method requires 146 ms/video. This speed is strictly faster than recent models such as LaSA (157 ms/video) and ME-ST (306 ms/video), while remaining highly competitive with DeST (114 ms/video). 
Notably, compared to the baseline (i.e., the architecture without MASF, AADL, and FACM), the integration of our frequency-selective modules introduces a modest overhead of 23 ms/video. This efficiency is primarily attributed to the inherently low time complexity of the FFT, which operates at $\mathcal{O}(T \log T)$. Consequently, the latency incurred by our spectral operations is substantially lower than that of stacking complex self-attention mechanisms or deep GCN operators. Given the significant performance improvements---such as the +4.8\% F1@50 absolute gain on the PKU-MMD (X-view) benchmark---this slight increase in computational cost is well-justified, demonstrating a highly favorable accuracy-efficiency trade-off.

\subsection{Robustness to Input Perturbations}
\label{app:compare:robustness}

\begin{table*}[tb]
    \footnotesize
    \centering
    \caption{Robustness evaluation under random Gaussian noise and joint occlusion on PKU-MMD v2 (X-sub) dataset. For each method, the first row reports the absolute performance, and the second row indicates the relative performance drop compared with the clean setting.}
    \setlength{\tabcolsep}{3pt}
    \begin{tabular}{c|ccccc|ccccc|ccccc}
        \toprule
        {Noise} & \multicolumn{5}{c|}{0.3x Gaussian Noise} & \multicolumn{5}{c|}{0.5x Gaussian Noise} & \multicolumn{5}{c}{30\% Joint Occlusion} \\
        \cmidrule(lr){2-6} \cmidrule(lr){7-11} \cmidrule(lr){12-16}
        
        Metric & Acc & Edit & \multicolumn{3}{c}{F1@\{10, 25, 50\}} & Acc & Edit & \multicolumn{3}{c}{F1@\{10, 25, 50\}} & Acc & Edit & \multicolumn{3}{c}{F1@\{10, 25, 50\}} \\
        \midrule
        
        DeST& 66.8 & 65.2 & 71.0 & 67.4 & 54.3 & 60.5 & 58.6 & 64.1 & 59.7 & 45.4 & 67.3 & 66.5 & 71.7 & 67.8 & 53.5 \\
        & \scriptsize \textcolor{olive}{-4.94\%} & \scriptsize \textcolor{olive}{-5.95\%} & \scriptsize \textcolor{olive}{-4.64\%} & \scriptsize \textcolor{olive}{-5.12\%} & \scriptsize \textcolor{olive}{-7.48\%} & \scriptsize \textcolor{olive}{-13.98\%} & \scriptsize \textcolor{olive}{-15.44\%} & \scriptsize \textcolor{olive}{-13.93\%} & \scriptsize \textcolor{olive}{-15.93\%} & \scriptsize \textcolor{olive}{-22.59\%} & \scriptsize \textcolor{olive}{-4.24\%} & \scriptsize \textcolor{olive}{-4.11\%} & \scriptsize \textcolor{olive}{-3.71\%} & \scriptsize \textcolor{olive}{-4.57\%} & \scriptsize \textcolor{olive}{-8.80\%} \\
        
        LaSA& 71.3 & 71.5 & 77.3 & 72.9 & 60.8 & 65.5 & 64.7 & 69.9 & 65.3 & 51.6 & 70.2 & 70.4 & 75.4 & 71.5 & 58.5 \\
        & \scriptsize \textcolor{olive}{-3.00\%} & \scriptsize \textcolor{olive}{-2.60\%} & \scriptsize \textcolor{olive}{-1.29\%} & \scriptsize \textcolor{olive}{-2.54\%} & \scriptsize \textcolor{olive}{-4.40\%} & \scriptsize \textcolor{olive}{-10.91\%} & \scriptsize \textcolor{olive}{-11.92\%} & \scriptsize \textcolor{olive}{-10.67\%} & \scriptsize \textcolor{olive}{-12.70\%} & \scriptsize \textcolor{olive}{-18.84\%} & \scriptsize \textcolor{olive}{-4.50\%} & \scriptsize \textcolor{olive}{-4.11\%} & \scriptsize \textcolor{olive}{-3.73\%} & \scriptsize \textcolor{olive}{-4.41\%} & \scriptsize \textcolor{olive}{-8.01\%} \\
        
        \textbf{SpecScalpel} & \textbf{74.7} & \textbf{74.1} & \textbf{79.3} & \textbf{76.2} & \textbf{66.0} & \textbf{73.9} & \textbf{73.2} & \textbf{78.3} & \textbf{75.6} & \textbf{64.8} & \textbf{72.7} & \textbf{72.0} & \textbf{77.4} & \textbf{74.2} & \textbf{62.3} \\
        & \scriptsize \textcolor{olive}{\textbf{-0.92\%}} & \scriptsize \textcolor{olive}{\textbf{-0.53\%}} & \scriptsize \textcolor{olive}{\textbf{-0.50\%}} & \scriptsize \textcolor{olive}{\textbf{-0.78\%}} & \scriptsize \textcolor{olive}{\textbf{-0.90\%}} & \scriptsize \textcolor{olive}{\textbf{-1.98\%}} & \scriptsize \textcolor{olive}{\textbf{-1.72\%}} & \scriptsize \textcolor{olive}{\textbf{-1.74\%}} & \scriptsize \textcolor{olive}{\textbf{-1.56\%}} & \scriptsize \textcolor{olive}{\textbf{-2.69\%}} & \scriptsize \textcolor{olive}{\textbf{-3.56\%}} & \scriptsize \textcolor{olive}{\textbf{-3.32\%}} & \scriptsize \textcolor{olive}{\textbf{-2.86\%}} & \scriptsize \textcolor{olive}{\textbf{-3.38\%}} & \scriptsize \textcolor{olive}{\textbf{-6.44\%}} \\
        \bottomrule
    \end{tabular}
    \label{tab:robust}
\end{table*}

To further assess the robustness of SpecScalpel against noise perturbations, we compare its performance degradation with that of other methods under random Gaussian noise and joint occlusion, as summarized in Tab.~\ref{tab:robust}. Specifically, all models are first trained on the original clean datasets and then evaluated on test sets corrupted by the noises. Gaussian noise with standard deviations of 0.3 and 0.5 times the original data scale is applied, while the random occlusion masks 30\% of the joints. For consistent and fair comparison, the same random seeds are used across all methods to ensure identical noise patterns.

Tab.~\ref{tab:robust} reports both the absolute performance under noisy conditions and the relative performance drop compared with the clean setting. As observed, SpecScalpel maintains high performance even under severe noise, exhibiting only minor degradation compared with other methods. In particular, its performance remains nearly unaffected by Gaussian noise and is still stable under occlusion scenarios. This robustness arises from the discrepancy-guided frequency-selective filtering, which effectively suppresses irrelevant high-frequency components caused by skeleton jitter or occlusion noise. In summary, the introduced frequency-domain modeling not only enhances performance but also substantially improves the model’s robustness to noise interference.

\subsection{Robustness to Boundary Ambiguity}
\label{app:compare:robustnessBA}

\begin{table}[tb]
    \footnotesize
    \centering
    \caption{Robustness evaluation against noisy training boundaries on the PKU-MMD v2 (X-sub) dataset. For each method, the first row reports the absolute performance under random boundary jitter ($\delta\sim\mathcal{U}(-25,25)$ frames), and the second row indicates the relative performance drop compared to the clean training setting.}
    \setlength{\tabcolsep}{3pt}
    \begin{tabular}{c|ccccc}
        \toprule
        Noisy Labels & Acc & Edit & \multicolumn{3}{c}{F1@\{10, 25, 50\}}\\
        \midrule
        
        DeST& 65.9 & 61.8 & 68.5 & 63.2 & 48.9 \\
        & \scriptsize \textcolor{olive}{\textbf{-6.21\%}}& \scriptsize \textcolor{olive}{-10.76\%}& \scriptsize \textcolor{olive}{-8.09\%} & \scriptsize \textcolor{olive}{-10.94\%} & \scriptsize \textcolor{olive}{\textbf{-16.78\%}} \\
        
        LaSA& 68.1 & 65.1 & 70.8 & 65.7 & 51.2 \\
        & \scriptsize \textcolor{olive}{-7.37\%} & \scriptsize \textcolor{olive}{-11.37\%} & \scriptsize \textcolor{olive}{-9.64\%} & \scriptsize \textcolor{olive}{-12.17\%} & \scriptsize \textcolor{olive}{-19.47\%} \\

        Baseline& 67.1 & 64.9 & 71.4 & 65.3 & 51.3 \\
        & \scriptsize \textcolor{olive}{-8.78\%} & \scriptsize \textcolor{olive}{-11.04\%} & \scriptsize \textcolor{olive}{-8.71\%} & \scriptsize \textcolor{olive}{-12.40\%} & \scriptsize \textcolor{olive}{-20.21\%} \\
        
        \textbf{SpecScalpel} & \textbf{70.0} & \textbf{67.6} & \textbf{74.0} & \textbf{69.1} & \textbf{55.2} \\
        & \scriptsize \textcolor{olive}{\underline{-7.11\%}} & \scriptsize \textcolor{olive}{\textbf{-9.28\%}} & \scriptsize \textcolor{olive}{\textbf{-7.10\%}} & \scriptsize \textcolor{olive}{\textbf{-10.00\%}} & \scriptsize \textcolor{olive}{\underline{-17.07\%}} \\
        \bottomrule
    \end{tabular}
    \label{tab:robustBA}
\end{table}

To further evaluate the robustness of our framework against boundary ambiguity and annotation errors, we conduct experiments introducing noisy boundary labels during training. Specifically, we apply a random jitter to the ground-truth action boundaries, shifting the start and end frames of each segment by a random offset $\delta\sim\mathcal{U}(-25,25)$ frames. The models are trained on these corrupted labels and evaluated on the clean test set.

As shown in Table \ref{tab:robustBA}, SpecScalpel consistently maintains the highest absolute performance across all metrics despite severe boundary perturbations. Furthermore, it exhibits a more resilient relative performance drop compared to the Baseline and LaSA method. This demonstrates that although the Adjacent Action Discrepancy Loss (AADL) utilizes ground-truth boundaries to extract segments for discrepancy computation, it does not overfit to precise temporal localizations. Because the dominant spectral peaks of human motion are statistically stationary, AADL only requires the extracted segment to predominantly capture the target action's characteristics. Even if the segment is truncated or slightly contaminated by adjacent actions due to boundary jitter, the model still effectively learns to amplify the discrepancy between the principal spectral components. This confirms that our discrepancy-guided filtering captures intrinsic, action-specific spectral patterns rather than overfitting to boundary annotation artifacts.

\subsection{Robustness to Temporal Rescaling}
\label{app:compare:robustnessTR}

\begin{table*}[tb]
    \footnotesize
    \centering
    \caption{Robustness evaluation against temporal rescaling on the PKU-MMD v2 (X-sub) dataset. Models trained on the original temporal scale are evaluated on sequences rescaled by 0.7x and 1.5x. For each method, the first row reports the absolute performance, and the second row indicates the relative performance drop compared to the clean setting.}
    \setlength{\tabcolsep}{3pt}
    \begin{tabular}{c|ccccc|ccccc}
        \toprule
        {Temporal Rescaling} & \multicolumn{5}{c|}{0.7x Temporal Rescaling} & \multicolumn{5}{c}{1.5x Temporal Rescaling} \\
        \cmidrule(lr){2-6} \cmidrule(lr){7-11}
        
        Metric & Acc & Edit & \multicolumn{3}{c}{F1@\{10, 25, 50\}} & Acc & Edit & \multicolumn{3}{c}{F1@\{10, 25, 50\}} \\
        \midrule
        
        DeST& 69.6 & 66.1 & 73.1 & 70.3 & 57.6 & 62.5 & 61.0 & 65.4 & 60.6 & 48.1 \\
        & \scriptsize \textcolor{olive}{-1.00\%} & \scriptsize \textcolor{olive}{-4.67\%} & \scriptsize \textcolor{olive}{-1.88\%} & \scriptsize \textcolor{olive}{-0.97\%} & \scriptsize \textcolor{olive}{-1.83\%} & \scriptsize \textcolor{olive}{\textbf{-11.16\%}} & \scriptsize \textcolor{olive}{-11.98\%} & \scriptsize \textcolor{olive}{\textbf{-12.20\%}} & \scriptsize \textcolor{olive}{\textbf{-14.68\%}} & \scriptsize \textcolor{olive}{\textbf{-18.11\%}} \\
        
        LaSA& 72.9 & 71.3 & 77.2 & 74.2 & 62.4 & 64.3 & 64.3 & 67.9 & 63.4 & 50.3 \\
        & \scriptsize \textcolor{olive}{-0.82\%} & \scriptsize \textcolor{olive}{-2.86\%} & \scriptsize \textcolor{olive}{-1.40\%} & \scriptsize \textcolor{olive}{-0.80\%} & \scriptsize \textcolor{olive}{-1.89\%} & \scriptsize \textcolor{olive}{-12.52\%} & \scriptsize \textcolor{olive}{-12.47\%} & \scriptsize \textcolor{olive}{-13.24\%} & \scriptsize \textcolor{olive}{-15.24\%} & \scriptsize \textcolor{olive}{-20.88\%} \\

        Baseline & 72.6 & 71.0 & 77.0 & 73.3 & 62.2 & 64.0 & 64.0 & 67.2 & 61.8 & 50.4 \\
        & \scriptsize \textcolor{olive}{-1.33\%} & \scriptsize \textcolor{olive}{-2.74\%} & \scriptsize \textcolor{olive}{-1.53\%} & \scriptsize \textcolor{olive}{-1.70\%} & \scriptsize \textcolor{olive}{-3.34\%} & \scriptsize \textcolor{olive}{-13.03\%} & \scriptsize \textcolor{olive}{-12.26\%} & \scriptsize \textcolor{olive}{-14.02\%} & \scriptsize \textcolor{olive}{-17.10\%} & \scriptsize \textcolor{olive}{-21.58\%} \\
        
        \textbf{SpecScalpel} & \textbf{74.8} & \textbf{72.5} & \textbf{78.6} & \textbf{76.3} & \textbf{65.4} & \textbf{66.1} & \textbf{65.6} & \textbf{69.2} & \textbf{65.4} & \textbf{53.4} \\
        & \scriptsize \textcolor{olive}{\textbf{-0.79\%}} & \scriptsize \textcolor{olive}{\textbf{-2.65\%}} & \scriptsize \textcolor{olive}{\textbf{-1.37\%}} & \scriptsize \textcolor{olive}{\textbf{-0.65\%}} & \scriptsize \textcolor{olive}{\textbf{-1.80\%}} & \scriptsize \textcolor{olive}{\underline{-12.38\%}} & \scriptsize \textcolor{olive}{\textbf{-11.95\%}} & \scriptsize \textcolor{olive}{\underline{-13.17\%}} & \scriptsize \textcolor{olive}{\underline{-14.81\%}} & \scriptsize \textcolor{olive}{\underline{-19.76\%}} \\
        \bottomrule
    \end{tabular}
    \label{tab:robustTR}
\end{table*}

To evaluate the model's robustness against variations in frame rates and action execution speeds, we conduct a temporal rescaling experiment. Models trained on the original temporal scale are evaluated on test sequences rescaled by factors of 0.7x and 1.5x. 

As shown in Table \ref{tab:robustTR}, SpecScalpel consistently achieves the highest absolute performance across all metrics, outperforming prior state-of-the-art methods by a significant margin. Furthermore, its relative performance degradation is tightly bounded and remains highly competitive with the baselines. This demonstrates that our model effectively learns the intrinsic spectral distributions of motion dynamics rather than overfitting to static frequency bins. We attribute this strong temporal scale robustness to the dynamic weighting branch within the Multi-scale Adaptive Spectral Filter (MASF). By dynamically adjusting the filter weights conditioned on the specific characteristics of the input instance, this branch adaptively compensates for the spectral shifts caused by temporal rescaling, thereby mitigating the brittleness associated with fixed, absolute-position frequency filtering.

\section{More Ablation Studies}
\label{app:ablation}

This section presents extensive ablation studies to rigorously validate the design choices of Spectral Scalpel. We analyze the placement and necessity of our core frequency-domain modules (Sec.~\ref{app:ablation:frequency}), followed by detailed ablations on the design parameters of the Multi-scale Adaptive Spectral Filter (MASF) (Sec.~\ref{app:ablation:MASF}), the Adjacent Action Discrepancy Loss (AADL) (Sec.~\ref{app:ablation:AADL}), and the Frequency-Aware Channel Mixer (FACM) (Sec.~\ref{app:ablation:FACM}). Finally, we ablate other architectural hyperparameters (Sec.~\ref{app:ablation:other}).

\subsection{Ablation Studies on Frequency Modeling}
\label{app:ablation:frequency}

\textbf{The Position of the Frequency Module.}
To validate our architectural design, we conducted a crucial ablation study to determine the optimal placement of the frequency modeling module (MASF+AADL). The primary goal of our module is to analyze features that retain distinct temporal frequency characteristics. We experimented with three strategic positions: (1) at the very beginning, processing raw skeleton coordinates; (2) after the spatial GCN modeling but before temporal modeling (our proposed approach); and (3) after the temporal modeling.
The empirical results, as summarized in Tab.~\ref{tab:position}, unequivocally demonstrate the superiority of our chosen position. Placing the frequency module after spatial modeling achieves the best performance across all metrics. We attribute this success to the nature of the features at this stage: the GCN has already enriched the frame-level representations with crucial semantic information regarding body structure and joint coordination. Critically, since no inter-frame temporal modeling has occurred, the information for each frame remains independent, thus preserving the rich, high-fidelity frequency dynamics of the motion, making it the ideal point for frequency analysis.

Conversely, positioning the module at the very beginning leads to suboptimal results. While applying FFT to raw kinematic signals is plausible, these low-level features lack the semantic context provided by the GCN, limiting the effectiveness of the frequency analysis. The most significant performance degradation is observed when the module is placed after temporal modeling, with F1@50 dropping sharply to 60.6\%. This confirms our hypothesis that temporal aggregation models inherently smooth the feature sequences to create homogeneous representations for classification. This process erases the motion frequency details that our module aims to capture, rendering a subsequent FFT largely ineffective.
In conclusion, this ablation study provides evidence that the optimal position for frequency analysis is after spatial feature extraction but before temporal aggregation. This placement ensures that the module operates on features that are both semantically rich and dynamically intact.

\begin{table}[tb]
    \centering
    \footnotesize
    \centering
    \caption{Ablation study on the position of the frequency module on PKU-MMD v2 (X-sub) dataset.}
    \renewcommand{\arraystretch}{1.0}
    \setlength{\tabcolsep}{4pt}
    \begin{tabular}{c|ccccc}
        \toprule
        Position of Frequency Module & Acc & Edit & \multicolumn{3}{c}{F1@\{10, 25, 50\}} \\
        \midrule
        At the Very Beginning & 74.8& 74.3& 78.8& 75.9& 65.6
\\
        \textbf{After Spatial Modeling} & \textbf{75.4}& \textbf{74.5}& \textbf{79.7}& \textbf{76.8}& \textbf{66.6}
\\
        After Spatial Modeling & 70.6& 70.0& 75.4& 71.9& 60.6\\
        \bottomrule
    \end{tabular}
    \label{tab:position} 
\end{table}

\textbf{The Necessity of Frequency-Domain Operations.}
A core design choice of our model is the deliberate use of the frequency domain for critical operations. To quantitatively validate this, we conducted an ablation study on our key modules: MASF and AADL. We tested variants where these modules were forced to operate directly in the time domain by removing the FFT/iFFT transformations.
The empirical results, summarized in Tab.~\ref{tab:FFT}, unequivocally demonstrate the superiority of our approach. Our full model, with both modules operating in the frequency domain, achieves the highest performance. A significant performance degradation is observed when either module is reverted to the time domain. The most substantial drop occurs when both modules operate in the time domain, confirming that the benefits of frequency-domain processing are cumulative and essential.

These results provide strong empirical validation for our theoretical justifications. For the MASF module: Time-domain filtering is fundamentally flawed for this task. It suffers from parameter meaning drift, as a filter at a specific timestep (e.g., frame 10) processes entirely different semantic actions across various sequences, hindering generalization. Furthermore, time-domain filtering is equivalent to frequency-domain convolution (by the Convolution Theorem), which causes spectral smearing (an effect that blurs and contaminates the very characteristic frequency peaks we aim to enhance). In contrast, the frequency domain provides an absolute and invariant axis (e.g., 2 Hz always corresponds to a ``twice per second'' motion). Our Hadamard product in this space performs a clean, global feature sharpening, which is a far more stable and effective optimization problem than learning a massive, equivalent global convolution kernel in the time domain. For the AADL loss: The performance drop from AADL w/o FFT confirms that direct time-domain comparison is unreliable. Such comparisons are easily dominated by confounding variables like action duration, scale, or repetition. By projecting all signals onto the same absolute frequency axis $[0, f_s/2]$, the FFT provides a fair and normalized basis for comparison. This allows the AADL to ignore superficial variations and focus on the intrinsic dynamic properties of the motions, providing more precise and meaningful guidance for model optimization.
In conclusion, this ablation study provides definitive quantitative evidence that the frequency domain is not merely an alternative but a superior representational space for both enhancing action-specific features and measuring sequence-level discrepancies, validating its central role in our architecture.

\begin{table}[tb]
    \centering
    \footnotesize
    \centering
    \caption{Ablation study on the necessity of frequency-domain operations on PKU-MMD v2 (X-sub) dataset.}
    \renewcommand{\arraystretch}{1.0}
    \begin{tabular}{c|ccccc}
        \toprule
        FFT or Not & Acc & Edit & \multicolumn{3}{c}{F1@\{10, 25, 50\}} \\
        \midrule
        \textbf{Both use FFT} & \textbf{75.4}& \textbf{74.5}& \textbf{79.7}& \textbf{76.8}& \textbf{66.6}
\\
        MASF w/o FFT & 75.1& 73.4& 79.1& 76.2& 65.5
\\
        AADL w/o FFT & 74.7& 73.6& 78.9& 75.9& 65.4
\\
        Both w/o FFT & 74.2& 73.2& 78.5& 75.4& 64.9\\
        \bottomrule
    \end{tabular}
    \label{tab:FFT} 
\end{table}

\subsection{Ablation Studies on MASF}
\label{app:ablation:MASF}

\textbf{Ablation Study on Filter Dimension and Weighting Strategy in MASF.}
To evaluate the design of multi-head channel-shared filter and channel-wise weighting in the MASF, we conduct an ablation study as presented in Tab.~\ref{tab:MASF}. Our goal is to achieve adaptive, per-channel spectral filtering in a computationally efficient manner. The study compares our proposed design against several variants, including less efficient, brute-force approaches. Tab.~\ref{tab:MASF} confirms the superiority of our strategy. Our proposed model (row 1), which uses channel-shared base filters with channel-wise weighting, achieves a high performance with a minimal addition of only 0.01G FLOPs and 0.006M parameters. This establishes a strong baseline for performance and efficiency.

The core of our design involves learning a small set of $M$ filters that are channel-shared (each $H_m \in \mathbb{R}^{1 \times R_m \times V}$) and then achieving adaptive channel-wise filtering via a dynamic-static channel-wise combination. Channel-wise weighting is the natural and necessary mechanism to achieve this. When we replace it with temporal-wise weighting (row 2) within the same filter-shared framework, performance drops significantly. This is because temporal weighting fails to create the per-channel filter specificity that is central to our design. Combining both weighting schemes (row 3) offers no improvement, suggesting that the temporal modulation is redundant and potentially disrupts the cleanly filtered signals. Furthermore, we compared our efficient model to computationally expensive alternatives using large, channel-specific filters (each $H_m \in \mathbb{R}^{C \times R_m \times V}$). A variant with only temporal weighting (Filter-Specific + Temporal-Weight, row 4) merely matches our performance but at a drastically higher computational cost. Adding both weighting mechanisms (Filter-Specific + Both Weights, row 5) yields only a marginal gain while incurring a massive parameter increase (+0.588M). In conclusion, this study validates that our design, which combines channel-shared base filters with a lightweight channel-wise weighting mechanism, represents the most elegant and cost-effective strategy. It achieves performance comparable to or better than computationally intensive alternatives, striking an optimal balance between effectiveness and efficiency.

\begin{table*}[tb]
    \centering
    \footnotesize
    \centering
    \caption{Ablation study on filter dimension and weighting strategy in MASF on PKU-MMD v2 (X-sub) dataset.}
    \renewcommand{\arraystretch}{1.0}
    \begin{tabular}{c|ccccc|cc}
        \toprule
        FFT or Not & Acc$\uparrow$  & Edit$\uparrow$ & \multicolumn{3}{c|}{F1@\{10, 25, 50\}$\uparrow$} & FLOPs$\downarrow$ & Param.$\downarrow$\\
        \midrule
        \textbf{Filter-Shared + Channel-Weight}& 75.4& 74.5& 79.7& 76.8& 66.6& \textbf{0.01G}&
\textbf{0.006M}
\\
        Filter-Shared + Temporal-Weight& 74.7& 74.2& 78.7& 76.0& 65.7& 0.045G&
0.010M
\\
        Filter-Shared + Both Weights& 75.2& 74.3& 78.9& 76.0& 66.0& 0.045G&
0.261M
\\
        Filter-Specific + Temporal-Weight& 75.5& 74.7& 79.5& 76.8& 66.6& 0.045G&0.337M
\\
        Filter-Specific + Both Weights & \textbf{75.6}& \textbf{74.9}& \textbf{79.5}& \textbf{77.0}& \textbf{66.8}& 0.045G&0.588M\\
        \bottomrule
    \end{tabular}
    \label{tab:MASF} 
\end{table*}

\textbf{Effect of Weighted Fusion Strategies in MASF.}
We evaluate different fusion strategies for aggregating multi-scale filtered features in MASF, including dynamic, static, and average fusion, and their combinations (Tab.~\ref{tab:5}). All strategies perform comparably across metrics, each offering unique advantages. Notably, combinations such as static-dynamic fusion yield further gains due to their complementary nature. Meanwhile, it can be observed that average fusion's non-learnability yields weaker complementary advantages than dynamic/static methods. The static-dynamic strategy achieves the best overall trade-off between robustness and adaptiveness, and is adopted in our final model.

\begin{table}[tb]
    \footnotesize
    \centering
    \caption{Effect of weighted fusion methods on multi-scale filtered features on PKU-MMD v2 (X-sub) dataset.}
    \setlength{\tabcolsep}{4pt}
    \renewcommand{\arraystretch}{0.7}
    \begin{tabular}{ccc|ccccc}
        \toprule
        Dynamic & Static & Average & Acc & Edit & \multicolumn{3}{c}{F1@\{10, 25, 50\}} \\
        \midrule
        \Checkmark & \XSolidBrush & \XSolidBrush & 75.2 & 74.0 & 79.4 & 76.5 & 66.2 
\\
        \XSolidBrush & \Checkmark & \XSolidBrush & 74.6 & 73.9 & 78.8 & 75.9 & 66.4 
\\
        \XSolidBrush & \XSolidBrush & \Checkmark & 75.2 & 74.2 & 79.4 & 76.5 & 66.3 
\\
        \Checkmark & \XSolidBrush & \Checkmark & 75.4 & 74.2 & 78.6 & 76.2 & 66.3 
\\
        \XSolidBrush & \Checkmark & \Checkmark & 75.3 & 74.4 & 79.5 & 76.4 & 66.4 
\\
        \Checkmark & \Checkmark & \XSolidBrush & 75.4 & \textbf{74.5} & \textbf{79.7} & \textbf{76.8} & \textbf{66.6} 
\\
        \Checkmark & \Checkmark & \Checkmark & \textbf{75.5} & 74.5 & 79.6 & 76.7 & 66.5 \\
        \bottomrule
    \end{tabular}
    \label{tab:5}
\end{table}

\textbf{Effect of Feature Domain and Weight Generation Strategy in Dynamic Weighting.}
To evaluate the impact of feature domain and weight generation strategy in the dynamic channel-wise weighting module, we conduct an ablation study as presented in Tab.~\ref{tab:7}. Specifically, we compare two sources of features for weight generation: temporal features prior to FFT ($F_s$) and frequency features after FFT ($F_{f0}$). Additionally, we examine two methods for generating the weights: our proposed formulation in Eq.~(3) and the SE block-based strategy inspired by~\cite{DFFormer}, where features of shape $\mathbb{R}^{C \times T \times V}$ are globally pooled along the temporal and spatial dimensions and passed through a two-layer SE block to generate weights of shape $\mathbb{R}^{C \times M}$. Experimental results indicate that using frequency-domain features leads to inferior performance compared to temporal features. This is attributed to the conversion from complex-valued frequency representations to amplitude spectra, which introduces information loss and degrades the effectiveness of the learned weights. Regarding the generation strategy, the SE block-based method also performs worse than our proposed approach. This is likely because global pooling suppresses fine-grained spatio-temporal cues and channel mixing is less compatible with the desired independent channel-wise modulation. In contrast, Eq.~(3) enables dynamic weight generation that fully leverages temporal and spatial structures in a channel-independent manner. Based on these findings, we adopt temporal features and Eq.~(3) for dynamic weight generation in our final model.

\begin{table}[tb]
    \footnotesize
    \centering
    \caption{Effect of feature domain and weight generation in dynamic weighting on PKU-MMD v2 (X-sub) dataset.}
    \setlength{\tabcolsep}{4pt}
    \renewcommand{\arraystretch}{1.0}
    \begin{tabular}{cc|ccccc}
        \toprule
        Feature Domain & Weight Generator & Acc & Edit & \multicolumn{3}{c}{F1@\{10, 25, 50\}} \\
        \midrule  
        Temporal & SE Block  & 74.8 & 73.4 & 78.4 & 75.6 & 65.6 \\
        Temporal & Equation (3)  & \textbf{75.4} & \textbf{74.5} & 79.7 & \textbf{76.8} & \textbf{66.6} 
\\
        Frequency & SE Block  & 74.8 & 74.3 & 78.6 & 75.5 & 65.6 \\
        Frequency & Equation (3)  & 75.3 & 74.2 & \textbf{79.8} & 76.8 & 66.4 \\
        \bottomrule
    \end{tabular}
    \label{tab:7}
\end{table}

\textbf{Effect of Weight Granularity and Initialization in Static Weighting.}
We investigate the influence of weight granularity and initialization strategy in the static channel-wise weighting branch, as shown in Tab.~\ref{tab:8}. For granularity, we compare a global configuration with shared weights across channels ($\mathbb{R}^{M \times 1}$) and a channel-specific configuration ($\mathbb{R}^{M \times C}$). Results reveal that channel-wise weights consistently outperform the global counterpart, suggesting that finer granularity allows more precise modulation of channel responses. As for initialization, both zero initialization and random initialization yield comparable performance after training, indicating that the static weights are sufficiently optimized regardless of their starting values. Consequently, our final design employs channel-wise granularity with zero initialization, offering both simplicity and strong empirical performance.

\begin{table}[tb]
   \footnotesize
    \centering
    \caption{Effect of granularity and initialization in static weighting on PKU-MMD v2 (X-sub) dataset.}
    \setlength{\tabcolsep}{4pt}
    \renewcommand{\arraystretch}{1.0}
    \begin{tabular}{cc|ccccc}
        \toprule
        Weight Granularity & Initialization & Acc & Edit & \multicolumn{3}{c}{F1@\{10, 25, 50\}} \\
        \midrule
        (M, C) & Random & 75.4 & \textbf{74.6} & 79.6 & 76.6 & 66.6 \\
        (M, C) & Zero  & \textbf{75.4} & 74.5 & \textbf{79.7} & \textbf{76.8} & \textbf{66.6} \\
        (M, 1) & Random & 75.4 & 74.3 & 79.6 & 76.3 & 66.3 \\
        (M, 1) & Zero & 75.3 & 74.4 & 79.6 & 76.4 & 66.4 \\
        \bottomrule
    \end{tabular}
    \label{tab:8}
\end{table}

\textbf{Effect of Fusion Strategy for Dynamic and Static Branch Features.}
We conduct an ablation study to assess different strategies for fusing features from the dynamic and static branches, as summarized in Tab.~\ref{tab:9}. Four fusion methods are compared: direct averaging, learnable scalar weighting, learnable channel-wise weighting, and channel-wise concatenation followed by convolution. Among them, the concatenation + convolution approach achieves the best performance by effectively capturing inter-branch correlations. However, direct averaging also delivers competitive results with significantly lower computational cost and no additional parameters. In contrast, the two learnable weighting schemes underperform, possibly due to their interference with the individual gradient flows of each branch, thus hampering optimization. Taking both performance and efficiency into account, we adopt direct averaging as the final fusion strategy in our model.

\begin{table}[tb]
    \footnotesize
    \centering
    \caption{Effect of  fusion strategy for dynamic and static branch features on PKU-MMD v2 (X-sub) dataset.}
    \setlength{\tabcolsep}{4pt}
    \renewcommand{\arraystretch}{1.0}
    \begin{tabular}{l|ccccc}
        \toprule
       Fusion Method & Acc & Edit & \multicolumn{3}{c}{F1@\{10, 25, 50\}} \\
        \midrule  
        Average Fusion  & 75.4 & 74.5 & \textbf{79.7} & \textbf{76.8} & 66.6 
\\
        Learnable Scalar Weight  & 75.4 & 74.2 & 79.6 & 76.7 & 66.3 
\\
        Channel-wise Weight  & 75.4 & 74.4 & 79.3 & 76.3 & 66.4 
\\
        Channel Concat + Conv  & \textbf{75.5} & \textbf{74.6} & 79.7 & 76.5 & \textbf{66.7} \\
        \bottomrule
    \end{tabular}
    \label{tab:9}
\end{table}

\textbf{Effect of the Number of Filter Heads $M$.}
We examine the influence of the number of filter heads $M$ in the multi-scale multi-head spectral filtering module, with results presented in Tab.~\ref{tab:10}. As $M$ increases from 2 to 8, we observe a consistent improvement in performance. This can be attributed to the enhanced diversity of frequency-aware filtered representations, akin to the feature disentanglement observed in multi-head attention mechanisms. A higher number of heads also introduces a greater variety of filter scales, enabling more comprehensive spectral coverage. However, increasing $M$ also leads to higher computational cost and parameter overhead. Moreover, the marginal gain in performance diminishes as $M$ continues to grow. Considering the trade-off between model complexity and accuracy, we set the number of heads to $M = 4$ in our final configuration.

\begin{table}[tb]
   \footnotesize
    \centering
    \caption{Effect of different numbers of filter heads \(M\) on PKU-MMD v2 (X-sub) dataset.}
    \setlength{\tabcolsep}{4pt}
    \renewcommand{\arraystretch}{1.0}
    \begin{tabular}{c|ccccc}
        \toprule
        Number of filter heads & Acc & Edit & \multicolumn{3}{c}{F1@\{10, 25, 50\}} \\
        \midrule
        2 & 75.4 & 74.4& 79.5 & 76.6 & 66.4 
\\
        4  & 75.4 & 74.5 & 79.7 & 76.8 & 66.6 
\\
        6 & 75.7 & 74.7 & 80.0 & 77.2 & 66.7 
\\
        8 & \textbf{75.8} & \textbf{74.9} & \textbf{80.1} & \textbf{77.4} & \textbf{66.7} \\
        \bottomrule
    \end{tabular}
    \label{tab:10}
\end{table}

\textbf{Effect of Maximum Filter Length $R_{\text{max}}$ in Spectral Filtering.}
To investigate the impact of filter scale, we conduct an ablation study on the maximum filter length $R_{\text{max}}$, which defines the upper bound of filter resolutions or scales across heads. Results in Tab.~\ref{tab:11} show that increasing $R_{\text{max}}$ from 32 to 256 consistently improves model performance, as longer filter lengths enable finer frequency resolution and more granular spectral filtering. Although this enhancement comes with a slight increase in parameter count and computation, the benefit in spectral precision is notable. In our final model, we set $R_{\text{max}} = 64$ to balance accuracy and efficiency, while noting that higher values of $R_{\text{max}}$ may be considered in resource-permissive applications for further performance gains.

\begin{table}[tb]
    \footnotesize
    \centering
    \caption{Impact of maximum filter length \(R_{max}\) in spectral filtering on PKU-MMD v2 (X-sub) dataset.}
    \setlength{\tabcolsep}{4pt}
    \renewcommand{\arraystretch}{1.0}
    \begin{tabular}{c|ccccc}
        \toprule
       Maximum Filter Length& Acc & Edit & \multicolumn{3}{c}{F1@\{10, 25, 50\}} \\
        \midrule  
        32  & 75.3 & 74.2 & 79.5 & 76.7 & 66.5 
\\
        64  & 75.4 & 74.5 & 79.7 & 76.8 & 66.6 
\\
        128  & 75.7 & 74.3 & 79.8 & 76.8 & 66.8 
\\
        256  & \textbf{75.8} & \textbf{74.7} & \textbf{79.8} & \textbf{77.0} & \textbf{66.9} \\
        \bottomrule
    \end{tabular}
    \label{tab:11}
\end{table}

\textbf{Effect of Different Multi-Scale Strategies.}
We compare the effectiveness of two multi-scale strategies for spectral filtering: the linear scaling approach proposed in Eq.~(2) of our paper and the traditional exponential scaling strategy defined as $R_m = 2^{m + b}$, as shown in Tab.~\ref{tab:12}. Both strategies yield modest improvements over the non-multi-scale baseline under a fixed maximum filter length of 64. Meanwhile, the multi-scale approach also results in smaller filter lengths (below 64) at other scales, thereby reducing both parameters and computational costs. Notably, the linear multi-scale strategy defined by Eq.~(2) outperforms the exponential counterpart. We attribute this to its ability to reduce overlap between interpolated filter boundaries, which facilitates finer-grained and more discriminative channel-wise modulation. Based on these findings, we adopt the strategy in Eq.~(2) as the default design.

\begin{table}[tb]
   \footnotesize
    \centering
    \caption{Comparison of multi-scale strategies with fixed maximum filter length 64 on PKU-MMD v2 (X-sub) dataset.}
    \setlength{\tabcolsep}{4pt}
    \renewcommand{\arraystretch}{1.0}
    \begin{tabular}{c|ccccc}
        \toprule
        Multi-scale strategies& Acc & Edit & \multicolumn{3}{c}{F1@\{10, 25, 50\}} \\
        \midrule
        Single-scale & 75.3 & 74.5 & 79.5 & 76.6 & 66.4 \\
        Equation~(2)  & \textbf{75.4} & 74.5 & \textbf{79.7} & \textbf{76.8} & \textbf{66.6} \\
        $R_m = 2^{m + b}$ & 75.4 & \textbf{74.6} & 79.6 & 76.7 & 66.5 \\
        \bottomrule
    \end{tabular}
    \label{tab:12}
\end{table}

\subsection{Ablation Studies on AADL}
\label{app:ablation:AADL}

\textbf{Ablation Study on the Scope of Action Discrepancy in Adjacent Action Discrepancy Loss.}
To examine the effect of the computation scope in the discrepancy loss of different actions, we compare three variants: (i) Adjacent Action Discrepancy Loss (AADL), which computes spectral differences only between adjacent actions in each instance, as detailed in Sec.~3.3; (ii) Instance-scope Action Discrepancy Loss (IADL), which compares each action with all other non-identical-class actions within the same instance; and (iii) Batch-scope Action Discrepancy Loss (BADL), which further extends this comparison to all non-identical-class actions across the mini-batch. 

As shown in Tab.~\ref{tab:23}, IADL and BADL achieve slightly better performance than AADL, with BADL achieving the highest score due to the richest discrepancy supervision.
However, the performance gap is marginal. We attribute this to two reasons. First, although AADL only leverages adjacent actions, it already captures critical spectral transitions at temporal boundaries, which are most relevant for action segmentation. Since segmentation errors predominantly occur near these transitions, adjacent discrepancies provide highly targeted supervision to sharpen boundary modeling. 
Second, although IADL and BADL introduce more extensive comparisons, all three variants effectively guide the Multi-scale Adaptive Spectral Filter (MASF) to suppress shared frequency components and enhance discriminative ones. As MASF is a global filter with limited capacity, its representational power may already be saturated even under local discrepancy supervision.
Moreover, AADL is significantly more efficient, as it only compares adjacent actions. In contrast, IADL and BADL require pairwise comparisons among all non-identical-class actions, resulting in a quadratic increase in computation with respect to the number of action segments---especially for BADL, where comparisons span the entire batch.  
Therefore, we adopt AADL as the default setting, offering the best trade-off between performance and efficiency.

\begin{table}[tb]
    \centering
    \footnotesize
    \centering
    \caption{Impact of action discrepancy scope in adjacent action discrepancy loss on PKU-MMD v2 (X-sub) dataset. AADL is Adjacent Action Discrepancy Loss, IADL is Instance-scope Action Discrepancy Loss, and BADL is Batch-scope Action Discrepancy Loss.}
    \renewcommand{\arraystretch}{1.0}
    \setlength{\tabcolsep}{4pt} 
    \begin{tabular}{c|ccccc}
        \toprule
        Discrepancy Calculation Scope & Acc & Edit & \multicolumn{3}{c}{F1@\{10, 25, 50\}} \\
        \midrule
        AADL & 75.4 & 74.5 & 79.7 & 76.8 & 66.6 
\\
        IADL & 75.5 & \textbf{74.6}& 79.7& 76.9& 66.6
\\
        BADL & \textbf{75.7} & 74.5& \textbf{79.9}& \textbf{77.1}& \textbf{66.7}\\
        \bottomrule
    \end{tabular}
    \label{tab:23} 
\end{table}

\textbf{Amplitude vs. Complex in AADL Computation.}
In our AADL, the spectral discrepancy between adjacent actions is computed using strictly their amplitude spectra, deliberately discarding phase information. To validate this design choice, we compare our amplitude-only approach against calculating the discrepancy using the full complex spectrum (which inherently includes both amplitude and phase). As shown in Table \ref{tab:Ampli}, incorporating phase information yields negligible performance variations while effectively doubling the computational overhead of the discrepancy calculation.
We attribute this to the crucial property of temporal shift-invariance. The discriminative characteristics of distinct actions are predominantly captured by their spectral energy distributions (amplitude). In contrast, phase primarily encodes non-discriminative temporal offsets or exact start times---an action retains its core semantic identity regardless of the specific phase at which the segment begins. Forcing the model to differentiate based on phase could inadvertently cause overfitting to specific temporal alignments and introduce unnecessary computational redundancy. Therefore, our intentional adoption of a phase-invariant amplitude spectrum ensures that AADL remains both robust to temporal shifts and highly efficient during training.

\begin{table}[tb]
    \footnotesize
    \centering
    \caption{Ablation on spectral representations (Amplitude vs. Complex) for computing the AADL on the PKU-MMD v2 (X-sub) dataset.}
    \setlength{\tabcolsep}{4pt}
    \renewcommand{\arraystretch}{1.0}
    \begin{tabular}{c|ccccc}
        \toprule
        Spectral Representation & Acc & Edit & \multicolumn{3}{c}{F1@\{10, 25, 50\}} \\
        \midrule  
        Amplitude Only& 75.4& \textbf{74.5}& \textbf{79.7}& \textbf{76.8}& \textbf{66.6}
\\
        Complex & \textbf{75.5}& 74.4& 79.6& 76.8& 66.6\\
        \bottomrule
    \end{tabular}
    \label{tab:Ampli}      
\end{table}

\textbf{Effect of AADL Computation Strategies.}
We examine various formulations of AADL, including different distance metrics and loss functions (Tab.~\ref{tab:6}). A normalized L1 difference (i.e., mean absolute deviation) achieves the best performance and highest efficiency. 
All loss functions project differences in $(0, \infty)$ to monotonically decreasing values in $(0, 1)$, differing mainly in their slopes. As a result, the overall performance variation is minimal.
We adopt $-\log(\mathrm{tanh}(\cdot))$ for its stable convergence and consistent performance.

\begin{table}[tb] 
    \footnotesize
    \centering
    \caption{Effect of different AADL computation strategies on PKU-MMD v2 (X-sub) dataset.}
    \setlength{\tabcolsep}{2.5pt}
    \renewcommand{\arraystretch}{0.9}
    \begin{tabular}{cc|ccccc}
        \toprule
        Discrepancy Metric & Loss Form & Acc & Edit & \multicolumn{3}{c}{F1@\{10, 25, 50\}} \\
        \midrule
        Normalized L1 & -log(tanh($\cdot$)) & 75.4 & \textbf{74.5} & \textbf{79.7} & \textbf{76.8} & \textbf{66.6} 
\\
        Normalized L2 & -log(tanh($\cdot$)) & 75.2 & 74.0 & 79.2 & 76.2 & 66.3 \\
        L-infinity & -log(tanh($\cdot$)) & 74.8 & 73.4 & 78.6 & 75.4 & 64.8 \\
        \midrule
        Normalized L1 & -log(B-sigmoid($\cdot$)) & 75.2 & 74.3 & 79.6 & 76.4 & 66.5 \\
        Normalized L1 & 1 - tanh($\cdot$) &\textbf{75.5} & 74.3 & 79.4 & 76.5 & 66.4 \\
        Normalized L1 & 1 - B-sigmoid($\cdot$) & 75.3 & 74.4 & 79.6 & 76.7 & 66.6 \\
        \bottomrule
    \end{tabular}
    \label{tab:6}
\end{table}

\textbf{Impact of Sampling Strategies for Action Spectral Features.}
To ensure the AADL can be computed meaningfully, all action segments must have a unified spectral length along the frequency dimension. Given that the raw spectral lengths of segments may vary---being either shorter or longer than the target length---we require a sampling strategy capable of both downsampling and upsampling. We investigate three such strategies: nearest-neighbor interpolation, linear interpolation, and adaptive average pooling, with results summarized in Tab.~\ref{tab:13}. Among them, nearest-neighbor interpolation performs the worst, as its downsampling process discards points, leading to potential aliasing, while its upsampling introduces staircase-like discontinuities that exaggerate adjacent action differences. Adaptive average pooling achieves moderate performance by smoothing upsampled sequences through averaging; however, it tends to excessively blur frequency details during downsampling, weakening inter-action discriminability. Linear interpolation achieves the best results, as it preserves critical spectral trends during both upsampling and downsampling while suppressing noise, thereby enabling discrepancy computation that better reflects the true physical differences between actions. Consequently, we adopt linear interpolation as our default sampling strategy.

\begin{table}[tb]
    \footnotesize
    \centering
    \caption{Impact of the sampling strategies for action spectral features on PKU-MMD v2 (X-sub) dataset.}
    \setlength{\tabcolsep}{4pt}
    \renewcommand{\arraystretch}{1.0}
    \begin{tabular}{c|ccccc}
        \toprule
        Sampling Strategies& Acc & Edit & \multicolumn{3}{c}{F1@\{10, 25, 50\}} \\
        \midrule  
        Linear Interpolation & 75.4 & \textbf{74.5} & \textbf{79.7} & \textbf{76.8} & \textbf{66.6} \\
        Nearest Neighbor & \textbf{75.6} & 73.6 & 79.4 & 76.3 & 65.9 \\
        AdaptiveAvgPool& 75.3 & 74.3 & 79.7 & 76.7 & 66.3 \\
        \bottomrule
    \end{tabular}
    \label{tab:13}      
\end{table}

\textbf{Effect of Fixed Sampling Length $S_f$ on Action Spectral Features.}
We further examine the influence of the fixed sampling length $S_f$ used for sampling action spectral features before computing AADL, as shown in Tab.~\ref{tab:14}. Fixing a unified spectral resolution is requisite for AADL to precisely align spectral axes for valid discrepancy comparisons. A length of $S_f = 16$ yields suboptimal performance due to insufficient frequency resolution, which leads to an inaccurate estimation of inter-action discrepancies. Increasing $S_f$ to 32 and 48 improves performance significantly, as these lengths better capture essential spectral patterns while maintaining computational tractability. Crucially, even for extremely short action segments, linear interpolation to this fixed length effectively preserves the critical spectral envelope without introducing disruptive artifacts. Beyond this range, additional gains are marginal, because most of the shortest action segments in the dataset have inherent spectral lengths near 32–48. Notably, increasing $S_f$ also leads to higher loss computation costs and slower convergence. Therefore, we select $S_f = 32$ as the default value, balancing accuracy and training efficiency.

\begin{table}[tb]
   \footnotesize
    \centering
    \caption{Effect of fixed sampling length \(S_{f}\) on action spectral features on PKU-MMD v2 (X-sub) dataset.}
    \setlength{\tabcolsep}{4pt}
    \renewcommand{\arraystretch}{1.0}
    \begin{tabular}{c|ccccc}
        \toprule
         Sampling Length& Acc & Edit & \multicolumn{3}{c}{F1@\{10, 25, 50\}} \\
        \midrule
        16& 75.3 & 73.6 & 78.9 & 76.1 & 66.2 
\\
        32& 75.4 & 74.5 & 79.7 & \textbf{76.8} & \textbf{66.6} 
\\
        48& \textbf{75.5} & \textbf{74.6} & \textbf{79.8} & 76.6 & 66.5 
\\
        64& 75.6 & 74.0 & 79.1 & 76.5 & 66.6 \\
        \bottomrule
    \end{tabular}
    \label{tab:14} 
\end{table}

\textbf{Impact of Discrepancy Scaling Factor $\alpha$ in Adjacent Action Discrepancy Loss.}
AADL maps the raw inter-action spectral discrepancies---ranging over $(0, \infty)$---to a bounded loss in $(0, 1)$ through a monotonically decreasing function. This transformation inherently introduces a saturation effect, where overly large discrepancies yield near-zero gradients, limiting further optimization. Rather than a drawback, this saturation is intentionally designed to act as a ``soft margin'' that bypasses easy, highly distinct sample pairs, thereby concentrating the training gradients on ambiguous and hard-to-distinguish adjacent action boundaries. To control this saturation threshold and the effective amplification of discrepancies, we conduct an ablation on the scaling factor $\alpha$, as presented in Tab.~\ref{tab:15}. We find that $\alpha = 100$ offers the best performance, effectively balancing sensitivity and gradient stability. In contrast, larger values ($\alpha = 200, 400$) lead to early saturation and hinder the model’s ability to distinguish subtle action differences. Conversely, smaller values ($\alpha = 50$) amplify discrepancies excessively, disrupting optimization stability and slightly degrading performance. Based on these findings, we adopt $\alpha = 100$ as the optimal scaling factor.

\begin{table}[tb]
    \footnotesize
    \centering
    \caption{Impact of discrepancy scaling factor \(\alpha\) in adjacent action discrepancy loss on PKU-MMD v2 (X-sub) dataset.}
    \setlength{\tabcolsep}{4pt}
    \renewcommand{\arraystretch}{1.0}
    \begin{tabular}{c|ccccc}
        \toprule
        Scaling Factor \(\alpha\) & Acc & Edit & \multicolumn{3}{c}{F1@\{10, 25, 50\}} \\
        \midrule  
        50 & 75.0 & \textbf{74.6} & 79.3 & 76.1 & 66.4 
\\
        100 & \textbf{75.4} & 74.5 & \textbf{79.7} & \textbf{76.8} & \textbf{66.6} 
\\
        200 & 75.3 & 73.9 & 79.2 & 76.3 & 66.0 
\\
        400 & 75.0 & 74.3 & 79.6 & 76.7 & 65.9 \\
        \bottomrule
    \end{tabular}
    \label{tab:15} 
\end{table}

\textbf{Effect of the Loss Weight $\lambda_3$ for Adjacent Action Discrepancy Loss.}
We evaluate the impact of the AADL loss weight $\lambda_3$ to determine its contribution to the overall training objective, as summarized in Tab.~\ref{tab:16}. When $\lambda_3 = 0.1$, performance deteriorates, suggesting that the AADL signal is too weak to influence the optimization trajectory, with gradients dominated by other loss terms. Increasing $\lambda_3$ to 1 significantly improves performance, indicating that the discrepancy constraint is now sufficiently enforced to guide the model toward learning adjacent action differences. Further increasing $\lambda_3$ beyond 1 shows negligible additional benefit, implying that the loss has reached a stable contribution threshold. Therefore, we set $\lambda_3 = 1$ as the default weight in our final configuration.

\begin{table}[tb]
   \footnotesize
    \centering
    \caption{Effects of the weight \(\lambda_3\) for adjacent action discrepancy loss on PKU-MMD v2 (X-sub) dataset.}
    \setlength{\tabcolsep}{4pt}
    \renewcommand{\arraystretch}{1.0}
    \begin{tabular}{c|ccccc}
        \toprule
         Weight \(\lambda_3\) & Acc & Edit & \multicolumn{3}{c}{F1@\{10, 25, 50\}} \\
        \midrule
        0.1& 75.4 & 74.4 & 79.6 & 76.7 & 66.3 
\\
        1.0& 75.4 & 74.5 & \textbf{79.7} & \textbf{76.8} & 66.6 
\\
        5.0& \textbf{75.5} & 74.5 & 79.6 & 76.7 & \textbf{66.7} 
\\
        10.0& 75.3 & \textbf{74.6} & 79.7 & 76.8 & 66.6 \\
        \bottomrule
    \end{tabular}
    \label{tab:16} 
\end{table}

\subsection{Ablation Studies on FACM}
\label{app:ablation:FACM}

\textbf{Impact of Frequency-Aware Channel Mixer and Stacking Depth.}
To analyze the role of FACM, we ablate the FFT operation, channel mixing strategy, and stacking depth, as presented in Tab.~\ref{tab:4}. Channel mixing without FFT yields minimal improvement, as similar mixing is already inherent in Linear Transformers. In contrast, applying FFT alone brings moderate gains by suppressing high-frequency noise. When combined with channel mixing, the frequency-aware operation facilitates learning robust time-frequency representations, resulting in improved performance. Furthermore, increasing the point-wise depth improves performance, with gains nearly saturating at a depth of 2.

\begin{table}[tb]
    \footnotesize
    \centering
    \caption{Impact of frequency-aware channel mixer and stacking depth on PKU-MMD v2 (X-sub) dataset.}
    \setlength{\tabcolsep}{4pt}
    \renewcommand{\arraystretch}{0.7}
    \begin{tabular}{ccc|ccccc}
        \toprule
        FFT & PConv & Depth  & Acc & Edit & \multicolumn{3}{c}{F1@\{10, 25, 50\}} \\
        \midrule
        \XSolidBrush & \XSolidBrush & - & 75.1 & 73.9 & 79.3 & 76.3 & 66.0 \\
        \Checkmark & \XSolidBrush & -  & 75.2 & 74.2 & 79.7 & 76.6 & 66.3 \\
        \XSolidBrush & \Checkmark & 2 & 75.1 & 74.1 & 79.3 & 76.4 & 66.1 \\
        \midrule
        \Checkmark & \Checkmark & 1 & 75.3 & 74.4 & 79.4 & 76.2 & 66.4 \\
        \Checkmark & \Checkmark & 2 & 75.4 & \textbf{74.5} & \textbf{79.7} & \textbf{76.8} & 66.6 \\
        \Checkmark & \Checkmark & 3 & \textbf{75.5} & 74.5& 79.7 & 76.7& \textbf{66.7} \\
        \bottomrule
    \end{tabular}
    \label{tab:4}
\end{table}

\textbf{Ablation on Complex-Valued Feature Processing in the Frequency-Aware Channel Mixer.}
Since the Frequency-Aware Channel Mixer (FACM) operates in the frequency domain, where features are represented as complex numbers, it is critical to consider how complex-valued features are handled during channel mixing. We conduct an ablation study on four design choices: (i) mixing only the real part, (ii) mixing only the imaginary part, (iii) applying shared weights to both parts, and (iv) using independent weights for real and imaginary components. Results are presented in Tab.~\ref{tab:17}. The first two strategies yield the worst performance, as separately processing either part breaks the physical consistency of complex-valued signals and results in information loss or distortion. The independent-weight strategy achieves moderate performance but is inferior due to its decoupling of real-imaginary interactions, increased model complexity, and reduced generalization capacity. The shared-weight configuration achieves the best performance by preserving the unified representation of complex features, reducing optimization difficulty, and aligning with signal processing priors regarding coherent frequency-domain transformations. Therefore, we adopt the shared-weight design as the final mixing scheme in FACM.

\begin{table}[tb]
    \footnotesize
    \centering
    \caption{Ablation on complex feature processing in frequency-aware channel mixer on PKU-MMD v2 (X-sub) dataset.}
    \setlength{\tabcolsep}{4pt}
    \renewcommand{\arraystretch}{1.0}
    \begin{tabular}{c|ccccc}
        \toprule
        Conv Strategy & Acc & Edit & \multicolumn{3}{c}{F1@\{10, 25, 50\}} \\
        \midrule
        Real-only & 74.5 & 73.5 & 78.7 & 75.9 & 64.5 
\\
        Imaginary-only & 74.7 & 73.5 & 77.9 & 74.7 & 63.5 
\\
        Shared Weights & \textbf{75.4} & \textbf{74.5} & \textbf{79.7} & \textbf{76.8} & \textbf{66.6} 
\\
        Independent Weights & 74.8 & 74.4 & 78.9 & 76.1 & 65.8 \\
        \bottomrule
    \end{tabular}
    \label{tab:17} 
\end{table}

\textbf{Ablation on Post-Processing Operations for Frequency-Aware Mixed Features.}
We further investigate the effect of three commonly used post-processing operations---ReLU activation, Batch Normalization, and Dropout---on the output of FACM before the residual connection, as shown in Tab.~\ref{tab:18}. Introducing ReLU results in performance degradation, likely because the time-domain signals reconstructed from FACM outputs contain both positive and negative oscillatory components (e.g., phase-related information), and ReLU truncates negative values, violating signal integrity. Similarly, Batch Normalization leads to degraded performance, possibly due to its disruption of amplitude patterns, which undermines the consistency of frequency-aware mixed features. In contrast, adding Dropout shows negligible impact on performance, indicating that the FACM produces robust frequency-domain features where moderate random feature point omission does not substantially impair spectral representations. Consequently, we do not employ any post-processing operations following FACM.

\begin{table}[tb]
   \footnotesize
    \centering
    \caption{Ablation on post-processing operations for frequency-aware mixed features on PKU-MMD v2 (X-sub) dataset.}
    \setlength{\tabcolsep}{4pt}
    \renewcommand{\arraystretch}{1.0}
    \begin{tabular}{c|ccccc}
        \toprule
        Post-Processing & Acc & Edit & \multicolumn{3}{c}{F1@\{10, 25, 50\}} \\
        \midrule
        Identity & 75.4 & \textbf{74.5} & \textbf{79.7} & \textbf{76.8} & \textbf{66.6} 
\\
        ReLU & 74.2 & 73.4 & 78.6 & 75.5 & 64.9 
\\
        BatchNorm & 75.1 & 74.3 & 78.9 & 76.3 & 65.4 
\\
        Dropout(0.3) & \textbf{75.6} & 74.5 & 79.6 & 76.5 & 66.5 \\
        \bottomrule
    \end{tabular}
    \label{tab:18} 
\end{table}

\textbf{Impact of Kernel Size Choices in Frequency-Aware Channel Mixer.}
By default, FACM employs point-wise convolution (i.e., kernel size = 1), enabling pure inter-channel interactions without modeling frequency-local dependencies. To explore whether broader receptive fields might benefit performance, we test larger kernel sizes in FACM, as reported in Tab.~\ref{tab:19}. Results show that increasing the kernel size consistently degrades performance. This can be attributed to the weak correlation and inherent physical independence among neighboring frequency bins; enforcing local interactions introduces spurious couplings, disturbs structured spectral representations, and injects non-physical noise. Thus, point-wise convolution is optimal for FACM, enabling efficient and semantically coherent cross-channel integration, while preserving orthogonality with the temporal modeling performed by the Linear Transformer branch.

\begin{table}[tb]
    \footnotesize
    \centering
    \caption{Impact of kernel size choices in frequency-aware channel mixer on PKU-MMD v2 (X-sub) dataset.}
    \setlength{\tabcolsep}{4pt}
    \renewcommand{\arraystretch}{1.0}
    \begin{tabular}{c|ccccc}
        \toprule
        Kernel Size & Acc & Edit & \multicolumn{3}{c}{F1@\{10, 25, 50\}} \\
        \midrule
        1 & \textbf{75.4} & \textbf{74.5} & \textbf{79.7} & \textbf{76.8} & \textbf{66.6} 
\\
        3 & 74.5 & 73.9 & 78.1 & 75.3 & 64.6 
\\
        5 & 73.8 & 74.2 & 78.5 & 75.7 & 64.6 
\\
        7 & 73.8 & 74.4 & 78.9 & 75.4 & 64.5 \\
        \bottomrule
    \end{tabular}
    \label{tab:19} 
\end{table}

\textbf{Comparison of FFT Normalization Strategies in Frequency-Aware Channel Mixer.}
We evaluate the effect of different FFT normalization schemes in the FACM module, as summarized in Tab.~\ref{tab:20}. The tested configurations include: (i) forward normalization (scaling by $1/S$), (ii) backward mode (no normalization), and (iii) orthogonal normalization (scaling by $1/\sqrt{S}$). Corresponding inverse FFT operations are adjusted accordingly to maintain scale consistency. Experimental results show minimal performance differences across all strategies, suggesting that the point-wise convolution layer effectively learns to compensate for scale variations induced by normalization. This observation aligns with theoretical expectations: when FFT and iFFT normalization are symmetrically paired, and the model includes learnable linear transformations, any scale inconsistency can be absorbed by the model weights. For simplicity, we adopt the backward normalization setting as the default.

\begin{table}[tb]
   \footnotesize
    \centering
    \caption{Comparing FFT normalization strategies in frequency-aware channel mixer on PKU-MMD v2 (X-sub) dataset.}
    \setlength{\tabcolsep}{4pt}
    \renewcommand{\arraystretch}{1.0}
    \begin{tabular}{c|ccccc}
        \toprule
        Normalization Mode & Acc & Edit & \multicolumn{3}{c}{F1@\{10, 25, 50\}} \\
        \midrule
        Forward (normalize by \(1/S\)) & \textbf{75.5} & \textbf{74.6} & \textbf{79.8} & 76.7 & 66.6 \\
        Backward (no normalization) & 75.4 & 74.5 & 79.7 & \textbf{76.8} & \textbf{66.6} \\
        Ortho (normalize by \(1/\sqrt{S}\)) & 75.3 & 74.6 & 79.5 & 76.4 & 66.5 \\
        \bottomrule
    \end{tabular}
    \label{tab:20} 
\end{table}

\subsection{Other Ablation Studies}
\label{app:ablation:other}

\textbf{Ablation on Temporal Block Depth in Temporal Modeling.}
Following the design of previous STAS methods, our temporal modeling module is constructed by stacking multiple temporal blocks to progressively capture long-range temporal dependencies across frames. We investigate the effect of varying the depth of temporal block stacking, as shown in Tab.~\ref{tab:21}. Experimental results demonstrate that increasing the number of temporal blocks consistently improves performance, indicating enhanced capacity to model complex temporal relationships. However, deeper networks incur higher computational and parameter overhead. Notably, performance gains begin to plateau beyond 10 layers, suggesting diminishing returns due to optimization saturation. To balance accuracy and efficiency, we adopt a depth of 10 temporal blocks as the default configuration in our framework.

\begin{table}[tb]
    \footnotesize
    \centering
    \caption{Ablation on temporal block depth in temporal modeling on PKU-MMD v2 (X-sub) dataset.}
    \setlength{\tabcolsep}{4pt}
    \renewcommand{\arraystretch}{1.0}
    \begin{tabular}{c|ccccc}
        \toprule
        block counts & Acc & Edit & \multicolumn{3}{c}{F1@\{10, 25, 50\}} \\
        \midrule
        6 & 72.1 & 71.0 & 76.7 & 73.4 & 62.1 
\\
        8 & 74.2 & 73.2 & 78.5 & 75.1 & 64.5 
\\
        10 & 75.4 & 74.5 & 79.7 & 76.8 & 66.6 
\\
        12 & 75.7 & \textbf{75.1} & \textbf{80.0} & 77.0 & 66.6 
\\
        14 & \textbf{76.0} & 75.0 & 79.6 & \textbf{77.1} & \textbf{66.8} \\
        \bottomrule
    \end{tabular}
    \label{tab:21} 
\end{table}

\textbf{Ablation on Stage Numbers in Classification and Boundary Refinement Branches.}
We further examine the effect of stage numbers in the refinement branches responsible for classification prediction and boundary regression, respectively. As reported in Tab.~\ref{tab:22}, for the classification branch, using one or two refinement stages based on Linear Transformers yields nearly optimal performance, with a single stage already sufficient to provide effective prediction refinement. Increasing the number of stages further leads to marginal or negative gains, likely due to overfitting. In contrast, the boundary refinement branch, which leverages dilated TCN layers, benefits from two refinement stages to reach peak performance, while deeper stacks again cause overfitting and degrade generalization. Considering both accuracy and computational cost, we employ a one-stage class prediction branch and a two-stage boundary regression branch in the final model.

\begin{table}[tb]
   \footnotesize
    \centering
    \caption{Ablation on the stage numbers of classification/boundary refinement branches on PKU-MMD v2 (X-sub) dataset.}
    \setlength{\tabcolsep}{4pt}
    \renewcommand{\arraystretch}{1.0}
    \begin{tabular}{cc|ccccc}
        \toprule
        Class stages & Boundary stages & Acc & Edit & \multicolumn{3}{c}{F1@\{10, 25, 50\}} \\
        \midrule
        1 & 1 & \textbf{75.7} & 74.1 & 78.2 & 75.3 & 66.2 
\\
        1 & 2 & 75.4 & \textbf{74.5} & \textbf{79.7} & 76.8 & 66.6 
\\
        2 & 2 & 75.3 & 74.5 & 79.6 & \textbf{76.9} & \textbf{66.7} 
\\
        2 & 3 & 74.8 & 74.0 & 79.6 & 75.9 & 66.0 
\\
        3 & 3 & 74.3& 74.5& 79.5& 76.2& 65.3\\
        \bottomrule
    \end{tabular}
    \label{tab:22} 
\end{table}

\section{Discussion}
\label{app:discussion}

This section provides a reflective discussion on the present work. We candidly address the model's current limitations (Sec.~\ref{app:discussion:limitations}), propose promising directions for future research (Sec.~\ref{app:discussion:future}), and consider the broader societal impacts and ethical implications of this technology (Sec.~\ref{app:discussion:broader}).

\subsection{Limitations}
\label{app:discussion:limitations}

Despite its compelling performance, Spectral Scalpel is not without limitations. As shown in Fig.~5, the model exhibits instances of misclassification and boundary offsets, particularly in challenging scenarios. Specifically, detailed in our per-class analysis (Sec.~\ref{app:compare:per-class}), the model's performance degrades for low-frequency or quasi-static actions, where motion is subtle. Furthermore, the global frequency spectrum proves insufficient for distinguishing between actions that are spectrally similar but semantically distinct. This includes temporally reversed pairs (e.g., ``Take off sth'' vs. ``Put on sth'') and actions that differ only in repetition or magnitude (e.g., ``Triple Lutz'' vs. ``Quadruple Lutz''). The primary challenge stems from the inherent limitations of using global FFT to capture spectra. Although FFT is effective and efficient, it is difficult to capture the full complexity of various human behaviors. In addition, unified learning and additional prior guidance for the core frequency representation of each action are also crucial. As the first systematic effort to introduce lightweight frequency analysis to Skeleton-based Temporal Action Segmentation (STAS), our work validates the approach's feasibility and effectiveness. However, capturing the intricate and varied motion patterns inherent in real-world behaviors remains a significant challenge due to the constraints of a global spectral representation and consistent learning for core action spectra.

\subsection{Future Work}
\label{app:discussion:future}

To address these limitations, we have identified two primary directions for future research.

First, to better capture quasi-static actions and resolve spectral ambiguities, we plan to transition from a purely global frequency analysis to a collaborative time-frequency analysis framework. The inherent limitation of the global FFT is its lack of temporal localization. Therefore, future work will explore adaptive local filtering techniques, such as the Short-Time Fourier Transform (STFT) or Wavelet Transforms. By integrating these methods, our goal is to develop a model that can capture localized, time-varying frequency patterns, enabling a more nuanced understanding of how motion spectra evolve throughout an action sequence.

Second, we aim to enhance the learning paradigm and guidance mechanisms. The current data-driven filtering approach, while powerful, can be further optimized. The existing spectral discrepancy loss focuses solely on maximizing inter-class distance, which may not guarantee the learning of a stable and consistent frequency representation for each action. To this end, we propose incorporating contrastive learning and prototype learning. This will encourage the model to learn a canonical and robust frequency prototype for each action class by contrasting positive and negative spectral patterns. Furthermore, the purely data-driven model is susceptible to local optima due to its high optimization randomness. To mitigate this, we will investigate methods for integrating action-specific frequency priors. By injecting domain knowledge into the learning process, we can guide the model towards more meaningful and generalizable solutions, thereby improving its overall robustness and performance.

\subsection{Broader Impacts}
\label{app:discussion:broader}

The proposed Spectral Scalpel framework for Skeleton-based Temporal Action Segmentation (STAS) offers promising societal benefits by enabling accurate, efficient, and privacy-conscious understanding of human motion, with applications in rehabilitation, elderly care, industrial monitoring, sports analysis, and human-computer interaction. By operating on skeletal data rather than raw video, the approach reduces privacy concerns and computational overhead, supporting broader and more ethical deployment. However, potential negative impacts include the misuse of STAS for surveillance, behavioral profiling risks when combined with other data sources, and fairness issues or safety hazards due to biased performance across demographic groups. Although this work is foundational, we recognize the importance of addressing these risks early. Mitigation strategies include promoting transparency in data use, auditing for bias, restricting deployment to consented and ethically reviewed settings, and implementing technical safeguards such as anonymization and access controls.
